\documentclass[fontsize=12 pt]{scrartcl}

\usepackage[table,dvipsnames]{xcolor}
\usepackage[draft,inline]{changes}
\usepackage{booktabs}
\setaddedmarkup{\textcolor{ForestGreen!100!black}{#1}}

\usepackage{setspace}
\usepackage[colorlinks=true, linkcolor=blue, citecolor=blue, urlcolor=blue]{hyperref}
\usepackage{amsmath,amssymb,amsfonts,amsthm,mathtools}

\usepackage{wrapfig}

\newcommand{\theoremref}[1]{\hyperref[#1]{Theorem~\ref*{#1}}}

\newcommand{\propositionref}[1]{\hyperref[#1]{Proposition~\ref*{#1}}}

\newcommand{\lemmaref}[1]{\hyperref[#1]{Lemma~\ref*{#1}}}

\usepackage[english]{babel}
\usepackage[T1]{fontenc}
\usepackage{graphics}
\usepackage{lmodern}
\usepackage{dsfont}
\usepackage{bbm}

\usepackage{dirtytalk}
\usepackage{amsmath}
\usepackage{tabularx,ragged2e}
\newcolumntype{Y}{>{\RaggedRight\arraybackslash\hspace{0pt}}X}
\newcolumntype{C}{>{\Centering\arraybackslash\hspace{0pt}}X}
\usepackage{ragged2e}
\usepackage{lipsum} 
\usepackage{array}

\newcolumntype{C}[1]{>{\centering\arraybackslash}m{#1}}
\usepackage{multirow}
\usepackage{graphicx}

\usepackage{booktabs}

\usepackage{color} 
\usepackage[defaultlines=2,all]{nowidow}
\usepackage{caption}
\usepackage[labelformat=simple]{subcaption}
\usepackage[section]{placeins}

\usepackage{blindtext}
\usepackage[style=authoryear-comp, dashed=false, natbib=true,maxcitenames=2 ,sortcites=no, maxbibnames=999, uniquelist=false,  uniquename=false]{biblatex}

\DeclareNameAlias{default}{last}

\makeatletter
\newcommand*{\rom}[1]{\expandafter\@slowromancap\romannumeral #1@}
\makeatother

\newcommand{\assumptionref}[1]{\hyperref[#1]{Assumption~\ref{#1}}}

\makeatother

\usepackage[top=2.5cm,bottom=2cm,right=2.5cm,left=3.5cm]{geometry}
\usepackage{fancyhdr}
\usepackage{eso-pic}    
\usepackage{tocloft}  
\newcommand{\blind}{0} 

\usepackage{lipsum} 
\usepackage{setspace}
\usepackage{authblk} 

\usepackage{setspace}
\usepackage{textcomp}  
\usepackage{authblk}    

\def\spacingset#1{\renewcommand{\baselinestretch}{#1}\small\normalsize}
\spacingset{1}
\usepackage{xurl}
\usepackage{bm}

\begin{document}

\if0\blind
{
    \title{Electricity Price Forecasting: Bridging Linear Models, Neural Networks and Online Learning}
    \author{
    Btissame El Mahtout\textsuperscript{a,b}\thanks{Corresponding author.
    E-mail address : btissame.el@tu-dortmund.de},\quad
   Florian Ziel\textsuperscript{a}\\
\textsuperscript{a}Chair of Data Science in Energy and Environment, University of Duisburg-Essen, Germany \\
\textsuperscript{b}Chair of Econometrics and Statistics, Technical University of Dortmund, Germany
    }
    \date{} 
    \maketitle
    
} \fi

\begin{center}
\textbf{Abstract}
\end{center}
\begin{abstract}

Precise day-ahead forecasts for electricity prices are crucial to ensure efficient portfolio management, support strategic decision-making for power plant operations, and enable effective battery optimization. However, developing an accurate prediction model is highly challenging in an uncertain and volatile market environment. For instance, although linear models generally exhibit competitive performance in predicting electricity prices with minimal computational requirements, they fail to capture relevant nonlinear relationships. Nonlinear models, on the other hand, can improve forecasting accuracy with a surge in computational costs. We employ a multivariate hybrid neural architecture that combines linear and nonlinear feed-forward neural structures. We then integrate this architecture with a novel partial online learning strategy that combines warm-starting with distinct hyperparameter configurations for each training stage. This strategy, which substantially reduces computational time, constitutes our key contribution. Unlike previous hybrid models, our {framework} also incorporates forecast combination through Bernstein Online Aggregation (BOA) to further improve forecasting accuracy. Compared with the considered state-of-the-art benchmark implementations in the reported setup, the proposed forecasting method significantly reduces computational cost while delivering better forecasting accuracy (11–12\% RMSE and 14–17\% MAE reductions). Our results are derived from a six-year forecasting study conducted on major European electricity markets.
\end{abstract}

	\spacingset{1.45} 

\thispagestyle{empty}

\newpage
\setcounter{page}{1}

\section{Introduction}

The day-ahead electricity market operates through daily auctions submitted before noon, where prices are determined for electricity to be delivered on the following day (\cite{neuhoff2016intraday}).  It is the most important market in electricity trading and the main instrument for decision making (\cite{m}). Accurate price forecasts support power plant operators in making optimal scheduling decisions for the following day. This is particularly important for units with technical constraints such as minimum downtime requirements and start-up costs. Precise forecasts are also essential for submitting competitive bids and, in turn, generating higher profits. They also play a key role in battery storage optimization by enabling operators to determine the most profitable charge–discharge strategy. Accurate forecasts are not only beneficial to supply-side decisions, but they are also crucial for demand-side management planning, such as designing cost-efficient electric vehicle charging strategies. In these decision-making settings, the time required to generate the forecast is equally important as its accuracy, since the market participants must use all available information while meeting strict deadlines. This underscores the importance of short-term predictive models with high precision within strict time constraints, which is the central objective of this paper.

The day-ahead forecast models can be classified into two main categories: Linear models and nonlinear ones. Linear models with an autoregressive structure, where the lags of the price are used as predictors, were popular (\cite{contreras2003arima}; \cite{zhou2006electricity}). Later, the inclusion of exogenous variables such as solar and load along the lags has improved accuracy (\cite{conejo2005forecasting}). More recently, shrinkage estimation methods such as least absolute shrinkage and
selection operator (LASSO) (\cite{tibshirani1996regression}) and Elastic Net (\cite{zou2005regularization}) have demonstrated strong predictive accuracy thanks to their ability to select relevant features (\cite{lear}; \cite{ziel2016forecasting}, \cite{ziel2018day}). 
Linear models are generally effective in predicting electricity prices and are highly valued for their interpretability and fast computation. However, due to their basic structure, they may still miss important nonlinear relationships, such as the interaction between variables which can lead to less accurate forecasts.

{Those interactions, on the other hand, could be captured by nonlinear models, such as decision-tree-based methods or neural networks}.
However, we will focus here on the deep learning methods, given their rapidly growing popularity in electricity price forecasting (\cite{a}). The multilayer perceptron (MLP) is one of the simple architectures that has already been employed to generate forecasts a few years back (\cite{szkuta2002electricity}; \cite{catalao2007short}). Later, a more sophisticated architecture, like a denoising autoencoder, originally popular in natural language processing, is employed for price forecasting (\cite{wang2016short}). Additionally, Recurrent neural network (RNN) is also widely employed due to its structure to capture temporal dependencies in time series data (\cite{dnn}; \cite{ugurlu2018electricity}; \cite{chinnathambi2018deep}). Note that the deep learning methods usually outperform the linear models in terms of accuracy but fails in terms of computational cost (\cite{a}). To address this trade-off, we build a hybrid model that combines the strengths of both approaches while following the standard framework checklist proposed by \textcite{a}.

 We {employ} a simple neural network architecture that combines input and output through two components. The first component is a generalized linear model that connects the input to the output directly, while the second is an MLP that connects the input to the output through one hidden layer with a nonlinear activation function. This architecture was first introduced by \textcite{google} to improve application recommendations.
{Thanks to this structure, the model explicitly captures the linear structure commonly present in electricity price forecasting. Consequently, the nonlinear component only needs to model the remaining nonlinear patterns, which in turn may restrict the effective parameter search region for the nonlinear network and potentially reduce estimation error.}
 In addition to this unique architecture, the model incorporates all key drivers of electricity price formation, including renewable generation, electricity demand, fuel and carbon prices, autoregressive dynamics, and calendar effects. This ensures that both fundamental and seasonal relationships are fully captured. It is worth emphasizing that the literature is rich in hybrid models that combine different algorithms to improve accuracy in electricity price forecasting (\cite{kim2015short}; \cite{ben2018forecasting}; \cite{bento2018bat}; \cite{bisoi2020short}; \cite{gomez2023electricity}; \cite{cu2025time} ). {However, to the best of our knowledge, the suggested hybrid model has not been explored in this context.}

To further improve accuracy, an ensemble method is implemented to average the individual forecast models. Combining the predictions of different models has been shown to enhance precision as compared to individual forecasts in many studies (\cite{raviv2015forecasting}, \cite{nowotarski2016combine}, \cite{maciejowska2020pca}). The arithmetic mean is the simplest averaging technique, such that equal weight is assigned to each individual forecast (\cite{stock2004combination}). In this work, we select fully adaptive Bernstein Online Aggregation (BOA) proposed by \textcite{wintenberger2017optimal} as an ensemble method. This approach assigns dynamic weights to individual models based on their recent performance. While fully adaptive BOA was previously used to predict the electricity load (\cite{de2023adaptive}; \cite{hirsch2024online}; \cite{himych2024adaptive}), { we are not aware of prior studies applying it to electricity price forecasting.}

In addition to improving accuracy, keeping the computational time at an efficient level is a key objective of this work. To this end, we implement a novel partial online learning approach designed to minimize the computational burden. This method {integrates warm starting with window-specific hyperparameter configurations to reduce the training time of the model.  Online learning algorithms were formerly employed in the electricity price literature, for instance  \textcite{hirsch2025online} used it to generate a probabilistic price forecast. {However, to the best of our knowledge, the proposed partial online learning approach has not yet been examined in the energy forecasting literature.}}

Finally, the performance of the proposed models is compared to two well-established and performing models in electricity price forecasting (\cite{a}), which are Lasso Estimated AutoRegressive (LEAR) (\cite{lear}) and the Deep Neural Network (DNN) (\cite{dnn}). {We also include a generalized additive model with online learning as an additional benchmark.} We implement a forecasting study using the suggested methodology for two major European day-ahead electricity markets: the German–Luxembourg and the Spanish markets, with the former being the largest in Europe. {The results show that the proposed models achieve higher forecasting accuracy than the considered benchmarks while exhibiting lower runtimes under the implemented forecasting pipelines.}

The main contributions of this work can be summarized as follows:
\begin{enumerate}

\item We introduce a new partial online learning approach suitable for both linear and neural network models, enabling fast and efficient learning.

\item We provide the first implementation of the proposed linear–nonlinear hybrid architecture for day-ahead electricity price forecasting.

\item We are the first to apply a fully adaptive BOA framework for model ensembling and forecast combination {in electricity price forecasting}.

\item We conduct an extensive multi-market empirical study using more than six years of hourly data, demonstrating the robustness and transferability of the proposed methodology across different market structures.

\item  The results show that the proposed models consistently outperform the benchmark models in terms of both accuracy and computational efficiency across two major European markets, highlighting the general applicability of the framework.

\end{enumerate}

Besides this introduction, the manuscript consists of five more sections. 
Section \ref{section2} provides an overview of the data. Section \ref{section3} presents a general framework of model design. The forecast study design and the evaluation measures are described in Section \ref{section4}, while results are presented in  Section \ref{section5}.
Section \ref{section6} summarizes the findings and provides an outlook for future work.

\section{Data}\label{section2}

The datasets used in this project are mainly extracted from the website  \href{https://www.entsoe.eu}{https://\allowbreak www.\allowbreak entsoe.\allowbreak eu} of the European Network of Transmission System Operators for Electricity (ENTSO-E), which is a platform that provides complete information on the European electricity market. 
We consider data from {2018-10-01} to 2025-01-14 for the two largest European electricity markets: German-Luxembourg\footnote{Note that in October 2018 the Austrian-German-Luxembourg price zone got split.} and Spain. Thus, each of the datasets contains 2298 days of hourly observations (55153 hours) of the main variables for the electricity market. {The total number of hourly observations is not divisible by 24 because the sample includes one additional hour due to the daylight saving time transition in October 2018. The subsequent daylight saving time transitions from 2019 onward offset each other, with one hour less in March and one hour more in October.
}

For each of the considered datasets, the objective variable is the hourly day-ahead electricity price, which we denote $P_{d,h}$ for the price at day $d$ and hour $h${,} where $h=0,\ldots, 23$. We also consider typical external data, particularly day-ahead forecasts from renewable energy sources, solar ($\text{Solar}_{d,h}$), wind onshore ($\text{WindOn}_{d,h}$) and wind offshore ($\text{WindOff}_{d,h}$),
as well as day-ahead load forecasts ($\text{Load}_{d,h}$). Note that wind offshore is only available on the German-Luxembourg dataset but not in the Spanish one, as Spain 
has no large operational offshore wind generation facilities (\cite{cerda2025present}). In addition to these fundamental variables, both datasets incorporate
four electricity market-related commodity price time series. These are closing prices of short-term commodity futures, of TTF natural gas price ($\text{NGas}_{d}$),
Brent oil price ($\text{Oil}_{d}$),
Rotterdam Coal price ($\text{Coal}_{d}$)
and carbon emission prices by the European Emission Allowances ($\text{EUA}_{d}$). The commodity prices are collected from \texttt{refinitiv datastream} (\cite{refinitiv_datastream}).

The fuel-related prices are the same for both datasets, since they are not country-specific. The missing values in these commodity variables are the only missing values in the German-Luxembourg dataset. Those missing are replaced with the last available values, which are in line with energy market literature and correspond to the martingale assumption on these price time series. In contrast, the Spanish dataset contains additional missing values in the renewable and load day-ahead variables. Those missing are imputed by predictions obtained from simple regressions of the day-ahead variable on its corresponding actual variables. Lastly, given the distortions caused by daylight saving time, the missing hour on the last Sunday of March is imputed by the means of the adjacent hours, and the duplicated values on the last Sunday in October are replaced by their average.

\begin{figure}[htb!]
\centering
\includegraphics[width=0.99\textwidth]{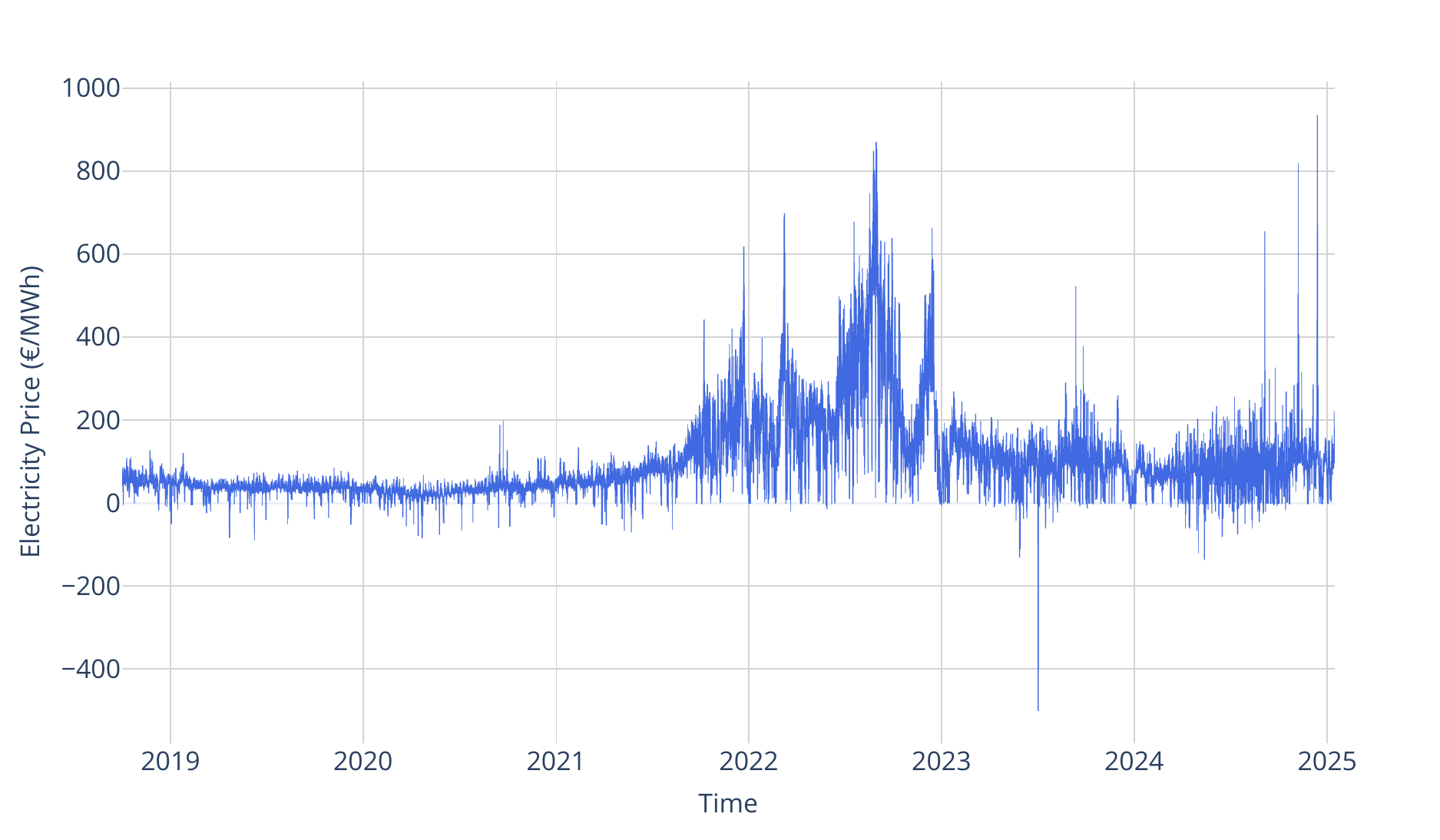}
\caption{{Time series of day-ahead electricity prices in Germany–Luxembourg
}}
\label{fig0}
\end{figure}

The time series in Figure \ref{fig0} provides a comprehensive overview of the changing dynamics of the electricity price in the German-Luxembourg bidding zone. Note that, in the main text, we focus on the data and result visualizations for this market, as it is the largest electricity market in Europe. The plots for the Spanish market are provided in the appendix. 
Examining this price series in Figure \ref{fig0}, one can see a high fluctuation, with a notable peak around August 2022, likely driven by the  energy crisis following the Russia-Ukraine conflict. Subsequently, prices gradually return to their typical interval, with some peaks and troughs that can be attributed to variations in demand and renewable generation. Another distinctive feature highlighted in Figure \ref{fig0}, which characterizes the German-Luxembourg market and is not common across all electricity markets, is the existence of negative prices. These can arise from subsidy schemes (renewable generators keep producing regardless of the market price) or production constraints (costs of keeping the power plant on could be smaller than shutting down and restarting it later).

\begin{figure}[htb!]
\centering
\includegraphics[width=0.90\textwidth]{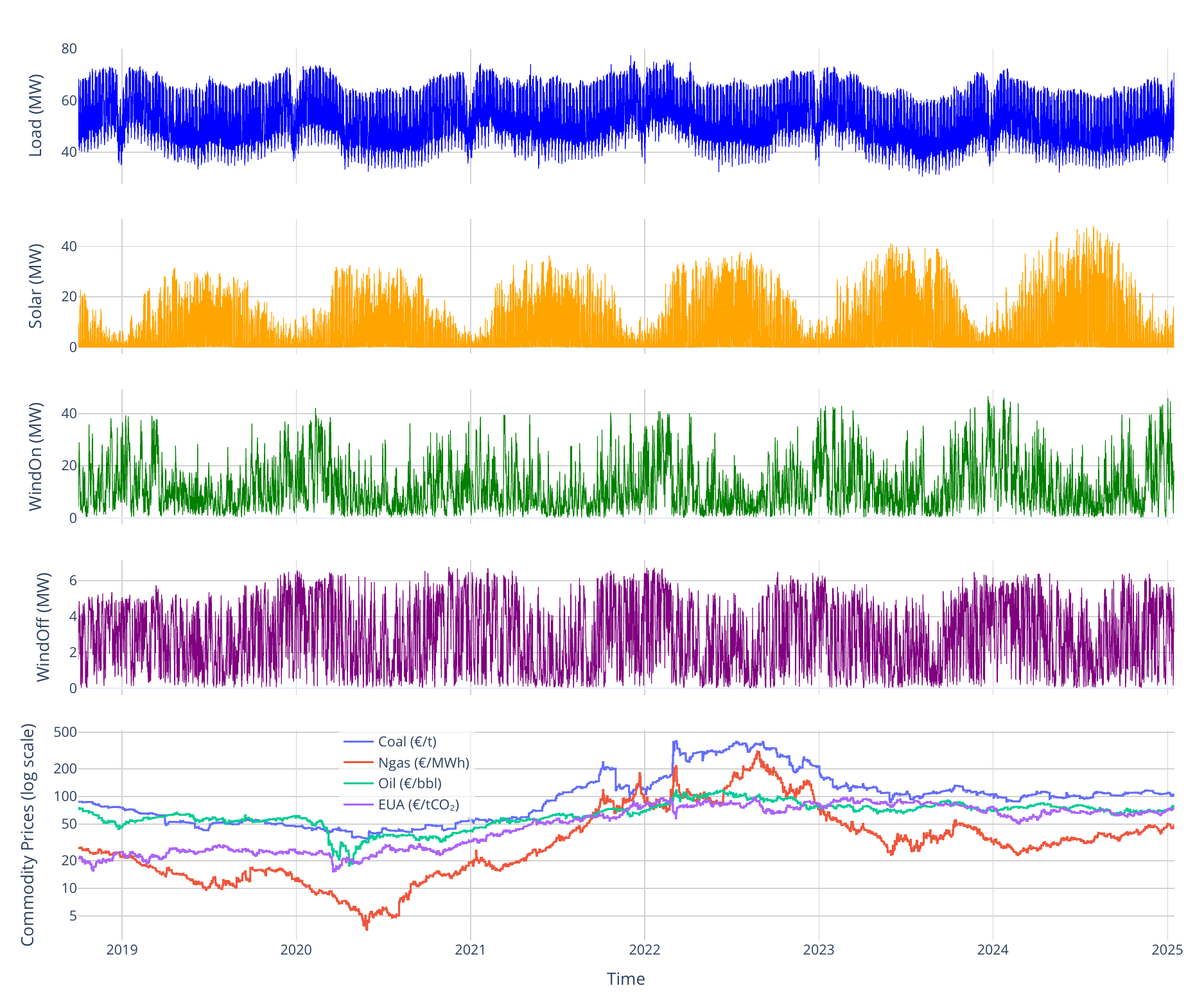}
\caption{Commodity prices and day-ahead forecasts of load and renewable generation in the German–Luxembourg electricity market
}
\label{fig03}
\end{figure}

Figure \ref{fig03} depicts the fundamentals and commodity variables employed as regressors to generate the forecast for the German-Luxembourg market. The bottom subplot highlights the surge in natural gas prices during the energy crisis. At the same time, the sharp increase in coal prices is attributed to the extensive use of coal-fired power during this crisis to compensate for the shortage of natural gas (\cite{huang2023research}). These developments likely contributed to the rise of electricity prices shown in Figure \ref{fig0}.

Fundamental variables in Figure \ref{fig03} show a clear annual seasonal pattern. Solar generation is low in winter and relatively high during the rest of the year. In contrast, electricity load is high during winter with a noticeable trough during the holiday season. In addition to yearly seasonality, electricity load exhibits weekly seasonality, which is not very apparent in the above Figure. The load is low during off-peak hours and weekends, which is usually associated with a decrease in electricity prices.

\section{Model Design}\label{section3}
Now we {present} and compare various models designed explicitly for day-ahead electricity price forecasting models. We examine two main configurations, linear and combined structure. Afterwards we explain the proposed ensembling method in detail.
\subsection{Linear Networks}
This architecture consists of an input layer and an output layer with 24 neurons, where each neuron represents the price of electricity at a specific hour. This setup is employed as the basis for two variants, referred to as the \textbf{Reduced Linear {(RLin)}} and \textbf{{Full Liner} {(FLin)}} models. The \textbf{{RLin}} model predicts the hourly electricity prices using regressors of the specific predicted hour. For instance, to predict hour zero (midnight to one o'clock in the morning), the model uses only the lagged prices and fundamentals corresponding to hour 0, along with time-invariant variables such as weekday dummies and commodity prices. The \textbf{{RLin}} model is expressed as follows:
\begin{align}
P_{d,h}&= \beta_{h,0}+\beta_{h,1} P_{d-1,h} + \underbrace{\beta_{h,2,1} P_{d-2,h} + \beta_{h,2,2} P_{d-7,h} }_\text{$=\boldsymbol{\beta}_{h,2}^\prime \mathbf{P}_{\cdot,h}$} + \beta_{h,3} P_{d-1,23} + \notag\\ 
&\quad \underbrace{\beta_{h,4,1}\text{Solar}_{d,h}+  \beta_{h,4,2} \text{WindOn}_{d,h} + \beta_{h,4,3}\text{WindOff}_{d,h}   + \beta_{h,4,4} \text{Load}_{d,h}}_\text{$=\boldsymbol{\beta}_{h,4}^\prime \mathbf{Fun}_{d,h} $}  + \notag\\ 
&\quad  \underbrace{\beta_{h,5,1}  \text{Oil}_{d-2} + \beta_{h,5,2} \text{Coal}_{d-2} +\beta_{h,5,3} \text{EUA}_{d-2}  + \beta_{h,5,4} \text{NGas}_{d-2}}_\text{$=\boldsymbol{\beta}_{h,5}^\prime \mathbf{Com}_{d-2}$}+ \notag\\ 
&\quad \underbrace{\beta_{h,6,1} \text{Mon}_{d} + \beta_{h,6,2} \text{Sat}_{d} +\beta_{h,6,3} \text{Sun}_{d}}_\text{$=\boldsymbol{\beta}_{h,6}^\prime \mathbf{Cal}_{d} $}+ \epsilon_{d,h} 
\label{eq1}
\end{align}

\begin{itemize}
    \item 
$\mathbf{P}_{\cdot,h} = \big( P_{d-2,h},\; P_{d-7,h}\big)^\prime \quad \text{(lag 2 and 7 of price)}, $ which includes price information at the same hour, $h$, 2 and 7 days ago.

\item $ \mathbf{Fun}_{d,h} = \big(\text{Solar}_{d,h},\; \text{WindOn}_{d,h},\; \text{WindOff}_{d,h},\; \text{Load}_{d,h}\big)^\prime \quad \text{(Fundamentals)},$ includes fundamental day-ahead forecast information at hour $h$ from load, wind and solar power

\item $\mathbf{Com}_{d-2}= \big(\text{Oil}_{d-2},\; \text{Coal}_{d-2},\; \text{EUA}_{d-2},\; \text{NGas}_{d-2}\big)^\prime \quad \text{(Commodities)}, $ represents a vector of the most recent commodity prices from natural gas, coal, oil and carbon emissions.
 
\item $\mathbf{Cal}_{d} = \big(\text{Monday}_{d},\; \text{Satuday}_{d},\; \text{Sunday}_{d}\big)^\prime \quad \text{(Calendar dummies)}, $ 

\item  $ \epsilon_{d,h}$ is the error term.

\end{itemize}


The coefficients $\beta_{d,j}$'s represent the model parameters to be estimated or trained. 
Commodity prices are modeled using information available on day $d-2$, since at the time of bidding (before noon on day $d-1$) the stock closing prices of commodities are not yet available.
{Lagged prices, fundamental variables, commodity prices, and calendar variables are selected as regressors as they are the most well-known predictors in the electricity price forecasting literature. While $P_{d-1,23}$ is included based on empirical evidence in \textcite{ziel2016forecasting} that the last hour of the previous day strongly influences early-day electricity prices.}

In addition to the\textbf{ {RLin}} model, we  consider a \textbf{{FLin}} model, which uses regressors from all hours rather than restricting them to the target predicted hour $h$. For example, when predicting the price for hour $h=0$, the model incorporates lagged prices and fundamentals from hours 0 through 23 as predictors, together 
with regressors that do not vary within the day. The following equation provides the \textbf{{FLin}} model specification:
{\small
\begin{align}
P_{d,h}&= \beta_{h,0}+\sum_{l=0}^{23} \beta_{h,l,1} P_{d-1,l} + \sum_{l=0}^{23} \underbrace{ \left(\beta_{h,l,2,1} P_{d-2,l} + \beta_{h,l,2,2} P_{d-7,l} \right)}_\text{$=\boldsymbol{\beta}_{h,l,2}^\prime \mathbf{P}_{\cdot,l}$} +
\notag \\ 
&\quad \sum_{l=0}^{23} \underbrace{\left(\beta_{h,l,4,1}\text{Solar}_{d,l}+  \beta_{h,l,4,2} \text{WindOn}_{d,l} + \beta_{h,l,4,3}\text{WindOff}_{d,l}   + \beta_{h,l,4,4} \text{Load}_{d,l}\right)}_\text{$=\boldsymbol{\beta}_{h,l,4}^\prime \mathbf{Fun}_{d,l} $}  + \notag\\ 
&\quad  \underbrace{\beta_{h,5,1}  \text{Oil}_{d-2} + \beta_{h,5,2} \text{Coal}_{d-2} +\beta_{h, 5,3} \text{EUA}_{d-2}  + \beta_{h,5,4} \text{NGas}_{d-2}}_\text{$=\boldsymbol{\beta}_{h,5}^\prime \mathbf{Com}_{d-2}$}+\notag \\ 
&\quad \underbrace{\beta_{h, 6,1} \text{Mon}_{d} + \beta_{h, 6,2} \text{Sat}_{d} +\beta_{h, 6,3} \text{Sun}_{d}}_\text{$=\boldsymbol{\beta}_{h,6}^\prime \mathbf{Cal}_{d} $}+ \nu_{d,h},
\label{eq2}
\end{align}
}
where $\nu_{d,h}$ is the error terms. The regressor $P_{d-1,23}$ is omitted from equation~\eqref{eq2}, as it is already included in the term $\sum_{l=0}^{23}\beta_{h,l,1} P_{d-1,l}$.

\subsection{nonlinear Networks and Combined structures}

The second main architecture in this project is a more elaborate network that, {besides the skip connection from the input to the output layer, employs a hidden layer equipped with the Leaky ReLU activation function to capture potential nonlinearities. This activation function is well-suited to electricity price data as it solves the dead-neuron problem that can arise in the presence of negative prices.} 

Two specifications of this architecture are considered: the \textbf{MLP with {RLin}} and the \textbf{MLP with {FLin}}. The \textbf{MLP with {RLin}} is the sum of the linear predictor of equation~\eqref{eq1} and a nonlinear part $\mathrm{NN}(\theta)$:
\begin{align}
P_{d,h} &=\; 
\beta_{h,0} 
+ \beta_{h,1} P_{d-1,h}+\mathbf{\beta}_{h,2}^\prime \mathbf{P}_{\cdot,h} + \beta_{h,3} P_{d-1,23} 
+ \mathbf{\beta}_{h,4}^\prime \mathbf{Fun}_{d,h} 
+ \notag \\
&\quad \mathbf{\beta}_{h,5}^\prime \mathbf{Com}_{d-2}  + \mathbf{\beta}_{h,6}^\prime \mathbf{Cal}_{d} 
+ 
\mathrm{NN}(\mathbf{\theta})
+\xi_{d,h},
\label{eq3}
\end{align}
\[
\mathrm{NN}(\mathbf{\theta})(\mathbf{x})
\;=\;
\mathbf{W}^{(2)}\,\sigma_{{\gamma}}
\bigl(
  \mathbf{W}^{(1)}\,\mathbf{x}
  \;+\;
  \mathbf{b}^{(1)}
\bigr)
\;+\;
\mathbf{b}^{(2)},
\;\;\;\;
\sigma_{{\gamma}}(z) \;=\;
\begin{cases}
z, & z \ge 0\\
{\gamma}\,z, & z < 0
\end{cases},
\]

where 
\(\mathbf{x} \in \mathbb{R}^{m}\) is the input vector with $m$ regressors, 
\(\mathbf{W}^{(1)}, \mathbf{W}^{(2)}, \mathbf{b}^{(1)}, \mathbf{b}^{(2)}\) 
are the trainable weight matrices and bias vectors, respectively. 
\(\mathbf{\theta}\) denotes the collection of all trainable parameters
and ${\gamma}=0.01$ {represents the default} leakage factor in the Leaky ReLU activation function.

\begin{figure}[htb!]
\centering
\includegraphics[width=0.95\textwidth]{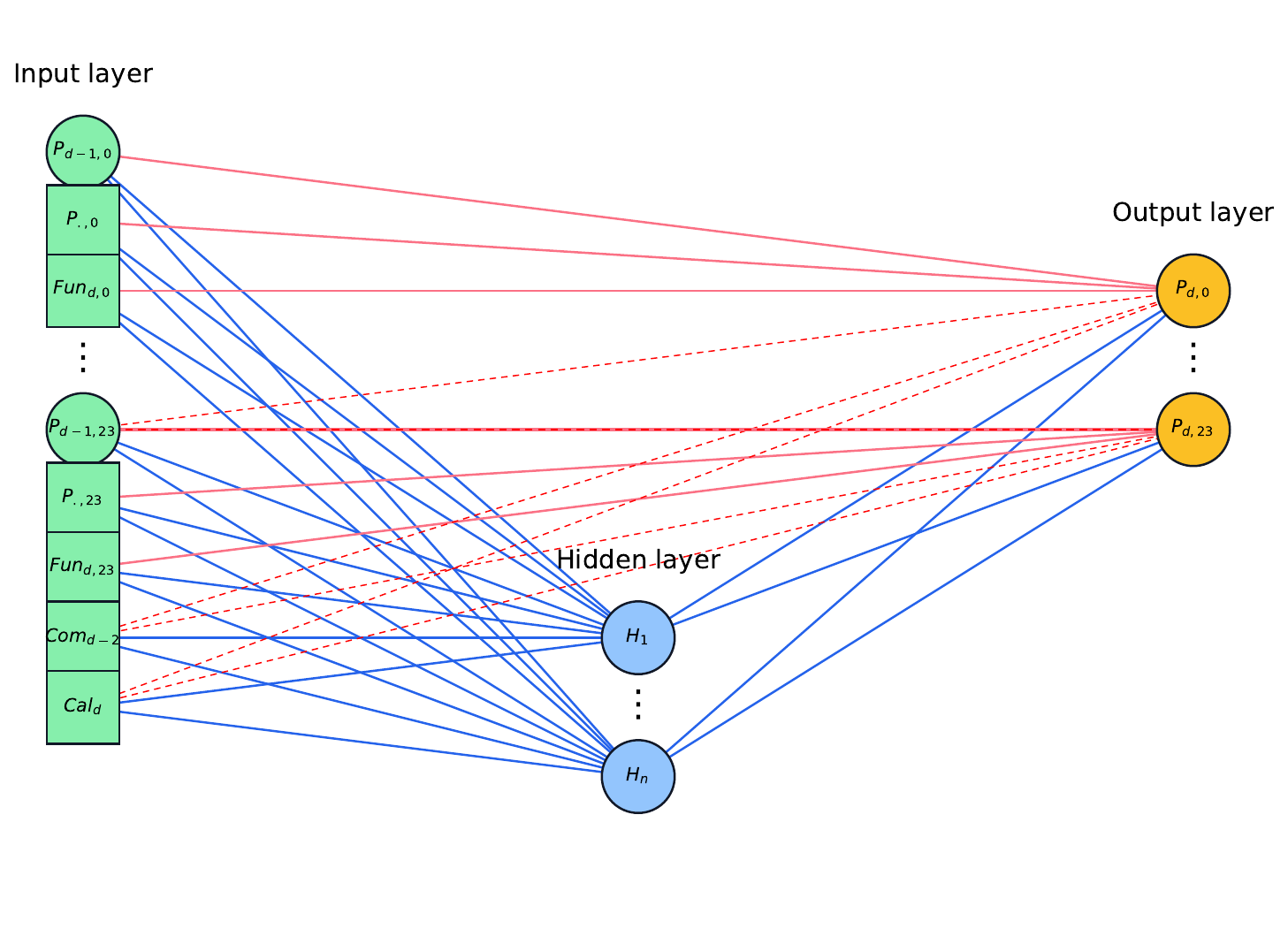}
\caption{{MLP with RLin model architecture for forecasting hourly electricity prices. The red connections represent the skip (linear) connections, while the blue connections correspond to the MLP component. Rectangular neurons {represent} the vector input.} }
\label{fig1}
\end{figure}

Figure \ref{fig1} illustrates the architecture of \textbf{MLP with {RLin}} Model. The green neurons represent the input layer. The hidden layer is defined by the blue $n$ neurons, where the number of neurons $n$ is a hyperparameter to be tuned. The output layer consists of 24 neurons, where each neuron provides the forecast for a specific hour of the day, is highlighted in orange. The blue edges linking the input and hidden layers correspond to the weight matrix $\mathbf{W}^{(1)}$, while the remaining blue connections are saved in $\mathbf{W}^{(2)}$. For simplicity, the bias vectors $\mathbf{b}^{(1)}$ and $\mathbf{b}^{(2)}$ are not shown in Figure~\ref{fig1}. Note that the rectangular neurons represent vectors rather than scalars. Thus, any connection from a rectangular node to another neuron does not represent a single weight, but a vector of weights. For instance, the blue edges connecting $\mathbf{Com}_{d-2}$ to the hidden neuron $\mathbf{H}_{1}$ illustrate four learnable weights, since $\mathbf{Com}_{d-2}$ is a 4-dimensional vector.

The red edges directly linking the input to the output, referred to as skip connection, represents the model’s linear predictor. The red solid lines connect to the variables that vary for each hour, whereas the dashed red lines correspond to variables that are constant throughout the day, such as fundamentals. However, an exception arises for the covariate $P_{d-1,23}$, which is constant across hours and also serves as the lagged price for predicting $P_{d,23} $. Accordingly, the skip connection for hours $P_{d,0}$ through $P_{d,22}$ includes 15 regressors, while the skip connection for $P_{d,23}$ contains only 14 covariates. It is worth noting that the linear relationships are not only learned through skip connection since the MLP component of the model can learn linear dependencies as well, which results in identifiability issues. {To address this, one could incorporate the orthogonalization cell proposed by \textcite{rugamer} to restore the interpretability of the linear components. In this study, however, we do not apply this procedure, as our primary focus is on predictive performance rather than on coefficient interpretability.}

\textbf{MLP with {FLin}} model shares the same hidden layer design as the \textbf{MLP with {RLin}} model, which is defined as $\mathrm{NN}(\mathbf{\theta})$ in equation~\eqref{eq3}. Yet differs in how the input-to-output connection is structured, as this linear component corresponds to that of the \textbf{{FLin}} model. The model equation is expressed in the following form:
\begin{align}
P_{d,h} &=\; 
\beta_{h,0} 
+ \sum_{l=0}^{23}\beta_{h,l,1} P_{d-1,l}+\sum_{l=0}^{23}\mathbf{\beta}_{h,l,2}^\prime \mathbf{P}_{\cdot,l} 
+ \sum_{l=0}^{23}\mathbf{\beta}_{h,l,4}^\prime \mathbf{Fun}_{d,l} 
+ \notag \\
&\quad \mathbf{\beta}_{h,5}^\prime \mathbf{Com}_{d-2}  + \mathbf{\beta}_{h,6}^\prime \mathbf{Cal}_{d} 
+ 
\mathrm{NN}(\mathbf{\theta})
+\mu_{d,h}.
\label{eq4}
\end{align}

Hence, the \textbf{MLP with {FLin}} model employs 175 regressors to forecast each hour. 


In addition to the four models introduced above, we develop three further variants. The first is the plain \textbf{MLP model}, which consists of an input layer, a hidden layer, and an output layer, without any skip connections such that the coefficients of the linear component in equation~\eqref{eq3} are set to zero ($\beta_{h,0} 
=\beta_{h,1} =\beta_{h,3} =
\mathbf{\beta}_{h,4}^\prime =\mathbf{\beta}_{h,5}^\prime =  \mathbf{\beta}_{h,6}^\prime =0$). The second and third are the \textbf{{RLin} with OLS} and the \textbf{MLP with {RLin} and OLS} models. These are structurally similar to the \textbf{{RLin}} and the \textbf{MLP with {RLin}} models, respectively, but differ in their weight initialization: the OLS-based variants use the coefficients of a simple linear model estimated via OLS as initial weights, whereas the non-OLS counterparts initialize weights randomly. 

It is important to highlight that each of the above models is implemented identically to predict electricity prices in the German-Luxembourg and Spain bidding zones. The only exception is that $\beta_{h,4,3}$ is set to zero in the Spanish case due to the absence of offshore wind generation data. 
{Furthermore, all the variables are standardized to enhance training stability. For each training window, standardization is performed by subtracting the mean and dividing by the standard deviation, employing the window-specific mean and standard deviation. The same statistics are subsequently used to standardize the observation to be forecasted with respect to the corresponding training window. This design ensures that the transformation relies only on information available within that window and that no future information is utilized, thereby preventing data leakage.}

\subsection{Ensembling}

Ensemble learning is a method where the forecasts of different models, referred to as experts, are combined with the goal of improving accuracy. Each of the models is multiplied by a specific weight and then aggregated to generate a single forecast such that the weights sum up to one:

$$\hat{P}_{d,h}^{\text{ens}} =  \sum_{k=1}^{K}{w_{k,d-1,h}} \hat{P}_{k,d,h}$$

where $\hat{P}_{d,h}^{\text{ens}}$is the ensemble forecast at day $d$ and hour $h$ , $\hat{P}_{k,d,h}$ is the forecast of model $k$, {$w_{k,d-1,h} $} is the corresponding weight, and $K$ is the number of models.

The simplest ensemble approach is to assign an equal weight to all models such that ${w_{k,d-1,h}} = \frac{1}{K}$. This weighting aggregation is employed, for example, by Random Forest (\cite{breiman2001random}), where the forecasts of different trees (experts) are averaged. However, the strength of  Random Forest lies not only in ensembling, but also in the construction of decorrelated trees and random feature selection \citep[Ch.~15]{hastie2009elements}.

A more advanced ensemble method is the fully adaptive Bernstein Online Aggregation (BOA) method proposed by \textcite{wintenberger2017optimal}. The weight of this approach varies among experts and between time steps for an individual expert. {The weights are computed as follows, using the implementation of the fully adaptive BOA presented in \textcite{berrisch2023crps}:}
 {
\begin{equation*}
\begin{aligned}
r_{k,d,h}
&=
\ell'(P_{d,h}, \hat{P}_{d,h}^{ens})
\left(
\hat{P}_{k,d,h} - \hat{P}_{d,h}^{ens}
\right)\\[4pt]
E_{k,d,h}
&=
\max\left(E_{k,d-1,h}, \lvert r_{k,d,h} \rvert \right)\\[4pt]
V_{k,d,h}
&=
V_{k,d-1,h} + r_{k,d,h}^{2}\\[4pt]
\eta_{k,d,h}
&=
\min\left(
\sqrt{\frac{-\log(w_{k,0,h})}{V_{k,d,h}}},
\frac{1}{2E_{k,d,h}}
\right)\\[4pt]
R_{k,d,h}
&=
R_{k,d-1,h}
+\frac12\Bigl(
r_{k,d,h}\bigl(1+\eta_{k,d,h}r_{k,d,h}\bigr)\Bigr)
+E_{k,d,h}\,\mathbf{1}\{2\eta_{k,d-1,h}r_{k,d,h} > 1\}
\\[4pt]
w_{k,d,h}
&=
\frac{
w_{k,0,h}\,\eta_{k,d,h}\,
\exp(-\eta_{k,d,h} R_{k,d,h})
}{
\sum_{j=1}^{K}
w_{j,0,h}\,\eta_{j,d,h}\,
\exp(-\eta_{j,d,h} R_{j,d,h})
},
\end{aligned}
\end{equation*}

where $w_{k,0,h}$ is the initial weight for expert $k$, which we set to 1/K. 
$\mathbf{1}\!\left\{.\right\}$ is the indicator function. $\lvert . \rvert $ is the absolute value. $\ell'(.)$ is the subgradiant of the loss. $\hat{P}_{d,h}^{ens}$ is the aggregate forecast and $\hat{P}_{k,d,h}$ is forecast of expert $k$, respectively, for day $d$ and hour $h$ using information up to day $d-1$.
} 
{The inclusion of $R_{k,d,h} $ inside the exponential when defining the weight, $w_{k,d,h}$,} ensures that experts who performed poorly in the past (up to day $d-1$) receive a smaller weight in day $d$, while larger weights are assigned to the best-performing experts in the past. Moreover, unlike classical ensemble methods such as Exponentially Weighted Average (EWA) (\cite{vovk1990aggregating}), fully adaptive BOA employed the second-order refinement that stabilises the learning process.

\section{Training and Evaluation Setup}\label{section4}

\subsection{Learning Algorithm}

In a normal rolling window forecast study, the forecast is generated following this mechanism:  data is divided into the test data with $P$ observations and training of size $D$, where $D=T-P$ and $T$ is the sample size. The model is fitted on observations $t=1$ to $t=D$, and the forecast is generated on observation $t=D+1$, which is the first observation in the test data. In the next step, the model is fitted on observations $t=2$ to $t=D+1$, and the forecast is generated on the second observation of the test data, $t=D+2$, and so on. 

The issue with this mechanism is that in each {training window}, the model is refitted on the last $D$ observations with only a single additional observation compared to the previous {training window}, which makes the procedure time-consuming. To reduce this computational burden, online learning is introduced such that instead of refitting {using} all $D$ observations, the weights are updated in each {training window} using the formerly estimated coefficients along the new observation. 
Online learning is applied in electricity price forecasting for both point and probabilistic predictions. For instance, \textcite{hirsch2024online} implements an algorithm that embeds online coordinate descent (OCD) or recursive least squares (RLS) into the Generalized Additive Models for Location, Scale and Shape (GAMLSS) framework to estimate regularized linear distributional models. Additionally, \textcite{yan2012online} employs an online/incremental Support Vector Regression (SVR) approach, which updates the model parameters as new data arrive, instead of retraining from scratch, to generate point forecasts of electricity prices.

\begin{figure}[h!]
\centering
\includegraphics[width=0.99\textwidth]{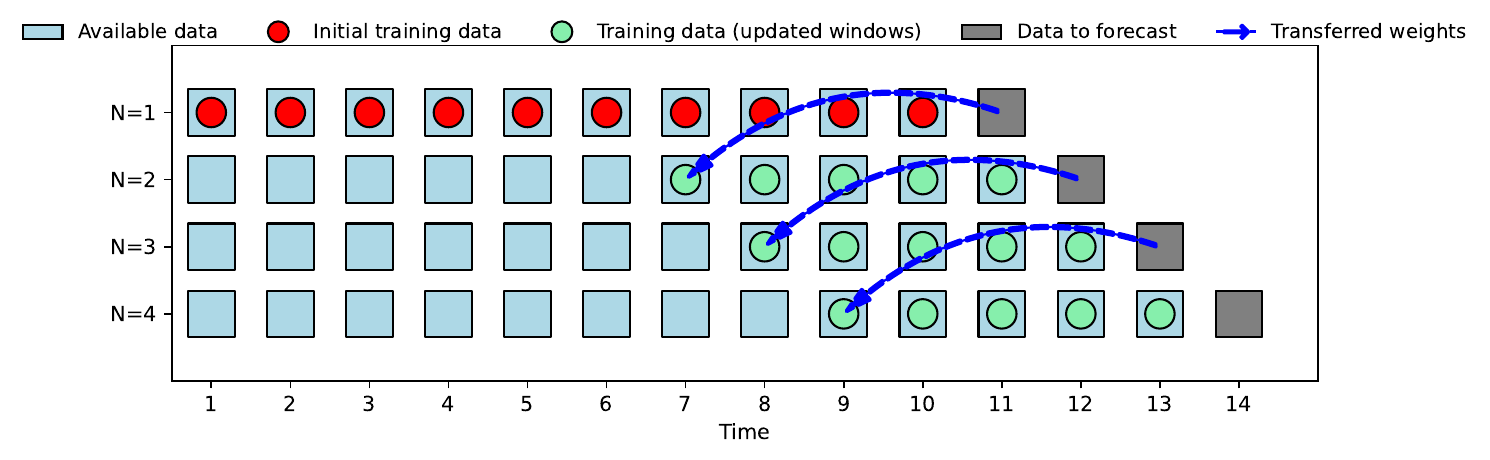}
\caption{Visualisation of the online learning algorithm for {$D_{\text{init}}= 10$ and $ D_{\text{up}}=5$}}
\label{fig2}
\end{figure}

This project, on the other hand, implements a partial online learning using a novel approach depicted in Figure  \ref{fig2}. In the first {training window}, the model is initialized randomly and fitted using 10 observations. The resulting weights are then employed to forecast observation 11. {Those weights are subsequently used to initialize the model in the second training window, where only the five most recent observations are used to refit the model.} Similarly, in {training window} 3, the optimal weights from {training window} 2 are carried forward, and the model is retrained on the latest five observations. This process continues till the final observation $T$ is forecasted. 
 
This approach is constructed on the assumption that the model requires a larger sample in the first {training window (referred to as the initial training window)} to learn the data structure, while in subsequent {training rounds (referred to as updated training windows)}, fewer observations suffice, since the carried-over weights already contain information from past data.
The initial 10 observations are referred to as the initial window size, {$ D_{\text{init}}$}, whereas the subsequent five observations used in each {updated training window} define the update window size {$ D_{\text{up}}$}. {$D_{\text{init}}= 10$ and $D_{\text{up}}= 5$ serve only as a simple illustration of how the online learning procedure functions. In practice, both $D_{\text{init}}$ and $D_{\text{up}}$ are two hyperparameters  that are tuned using the validation dataset.}

{The window size is not the only component for the proposed partial online learning to work effectively. Other elements are equally essential, as they also take different values in the initial training window than in the updated training window,  notably:} the number of epochs, which determines the number of times the model sees the full dataset, and the learning rate, which controls the magnitude of parameter updates during training. The number of epochs in the initial {training window} should generally be larger than the number of epochs in the updated windows. Since a higher number of epochs is associated with a larger number of weight updates, this ensures convergence toward a local minimum. However, for the subsequent {training window}, the model starts in the neighborhood of the minimum thanks to initialization weights from the previous {training window}, and then a small number of epochs is usually adequate to ensure convergence. A similar principle applies to the learning rate { and the penalty terms in the loss function defined in Equation \ref{eq:loss}. Note that since the window size is not the only component of the proposed online learning framework, there may be cases in which $D_{init}$ is smaller than $D_{up}$ (a configuration that is uncommon in standard online learning), yet the resulting approach remains faster than training without the proposed framework, due to the influence of the other hyperparameters.}

{The transfer of model parameters from one training round to the next, referred to as warm starting, is widely used in convex optimization problems. For example, \textcite{he2014practical} employ warm starting to incrementally update logistic regression models. In contrast, warm starting is rarely used in deep neural networks (\cite{ash2020warm}). However, it is important to emphasize that the novelty of the proposed partial online learning framework does not stem solely from the use of warm starting. Rather, it lies
in combining warm starting with window-specific hyperparameter configurations, where the initial and updated windows use hyperparameters tailored to their respective stages. Moreover, the term partial comes from the use of a couple of observations in the updated training rounds, $D_{up}>1$, not the new observation only, $D_{up}=1$, which is the standard in the online learning algorithm. Although setting $D_{up}=1$ is possible under the proposed learning approach, it is typically suboptimal in practice.}

\subsection{Study Design}\label{section4.2}

The dataset is split into training, validation, and test sets, where the test data comprises the last two years of observations from 2023-01-16 to 2025-01-14, and the validation set includes the two prior years of observations {(2021-01-16 to 2023-01-15)}. Training data, on the other hand, changes in every {training window}. For instance, during the hyperparameter tuning phase, the training set is the $ D_{\text{init}} $ observations prior to 2021-01-16, which marks the first observation in the validation dataset. The second training set then comprises the $ D_{\text{up}}$ observations preceding the second validation point, and so on.

{Each of the models passes through two main phases: hyperparameter tuning and evaluation. During the first stage, the best hyperparameters (e.g, number of neurons) are selected based on the performance on the validation dataset. In the later phase, the model with the selected hyperparameters is assessed on the test dataset, which remains entirely unseen during training and tuning. During the two stages, the models are recalibrated daily, such that the network weights are updated in each training window. While the daily refit of models is common (or even advisable) during the evaluation stage, it is not typically performed in the tuning phase. For instance, the benchmark model, DNN, is fitted only once during tuning. However, due to the unique structure of the proposed online learning, where the value of most hyperparameters varies depending on the initial and updated training window, we follow a daily recalibration even in the tuning phase. This approach allows us to identify hyperparameters that are optimal for the corresponding training window.}

{The hyperparameters to be tuned are} $D_{\text{init}}$  and $D_{\text{up}}$, along with the number of neurons in the hidden layer and initial and updated learning rate. In addition, the penalty terms $\lambda_1$ and $\lambda_2$ in the loss function (as defined in the following equation) are also subject to tuning:

\begin{equation}
L(\boldsymbol{\theta}) =
\ell(\mathbf{P}) +
\lambda_1 \|\boldsymbol{\theta}\|_2^2 +
\lambda_2 \|\boldsymbol{w}^{(\mathrm{out})}\|_1
\label{eq:loss}
\end{equation}
 where $\ell(\mathbf{P})$ is L1 loss, $\|\boldsymbol{\theta}\|_2$ is the Euclidean norm of the weight vector $\boldsymbol{\theta}$, and $\|\boldsymbol{w}^{(\mathrm{out})}\|_1$ is norm of the output-layer weights. The use of this regularized loss function is necessary to avoid overfitting, which is particularly important given the large number of regressors. The last hyperparameter to consider for tuning is solely employed by the two OLS-based models. {We denote this parameter by $\alpha$} and it determines the share of the OLS coefficient, $\boldsymbol{\hat{\beta}}_{\text{OLS}}$, to be used for initializing the skip connection of \textbf{{RLin} with OLS} and the \textbf{MLP with {RLin} and OLS models}. For example, if $\alpha=1$,  the initialization is given directly by $\boldsymbol{\hat{\beta}}_{\text{OLS}}$, while if $\alpha=2$, then models will use $2 \times \boldsymbol{\hat{\beta}}_{\text{OLS}}$ as initialisation.
 
The Tree-structured Parzen Estimator (TPE) optimization method, proposed by \textcite{TPE}, is employed to select the combination of optimal values of the above hyperparameters. TPE is a Bayesian optimization algorithm that selects candidate hyperparameters by modeling the distribution of good and bad trials from past evaluations to minimize the objective function. Unlike random search, which samples random values within the proposed interval in each trial, TPE employs previous trials to guide the search toward more promising regions of the hyperparameter space. We implement this hyperparameter optimization algorithm using the Python package Optuna (\cite{optuna}), where we set the number of trials to 500 and set the mean absolute error as an objective function. {The search space of the hyperparameters is illustrated in Table \ref{tab:hyperparameters} in the Appendix.
The use of 500 trials corresponds to evaluating 500 candidate sets of hyperparameters, aiming for a balance between achieving high forecast accuracy and keeping a moderate computational cost.}

On the other hand, there are other hyperparameters that we predetermine, such as the number of epochs, which we set to 60 for the initial {training round} and 10 in the subsequent {training rounds}, and the optimization algorithm, which we fix to Adaptive Moment Estimation (Adam) as it is simple and computationally efficient (\cite{adam}).

During hyperparameter tuning, 500 different combinations of hyperparameters are created for each of the 7 model classes discussed in section \ref{section3}. The combination that yields the smallest MAE is employed for fitting and generating the forecast in the test data. Additionally, for each model class $i$, where $i=1,2,...7$ we select the ten models that yield the best ensemble performance using the following iterative procedure:
\begin{itemize}
 \item Start with the model  $\mathcal{M}_1^i$  that achieves the lowest MAE among all 500 candidate models.
 \item Form pairwise BOA ensembles of  $\mathcal{M}_1^i$  with each of the remaining 499 models. Select  $\mathcal{M}_2^i$, the model that together with  $\mathcal{M}_1^i$  produces the lowest MAE.
 \item Next, form BOA ensembles of the current set $\{\mathcal{M}_1^i, \mathcal{M}_2^i\}$  with each of the remaining 498 models. Select  $\mathcal{M}_3^i$,  the model that yields the lowest MAE when added to the ensemble.
 \item Continue this forward step selection process until an ensemble of 10 models has been constructed.
 \end{itemize}

The resulting model for class $i$ will be {referred to as} {[Class Name] (BOA)}. For instance, the ensemble resulting from the \textbf{{RLin}} architecture will be referred to as \textbf{{RLin} (BOA)}. In addition to these seven class-specific ensembles, we also develop a general ensemble model, referred to as \textbf{BOA all}. This ensemble is built from the 500 models with the lowest MAE across all model classes. The same forward selection strategy is applied to identify the best combination of ten models.

{This forward selection strategy is not guaranteed to be optimal, since, for example, a  model that performs badly when selecting $\mathcal{M}_j^i$ may become highly informative when combined with a later selected model $\mathcal{M}_k^i$, outperforming the ensemble that includes $\mathcal{M}_j^i$, for $1 \le j < k \le 10$. However, this approach is computationally efficient, as it evaluates only 4456 ensemble combinations, compared to the $\binom{500}{10}$ possible combinations required by an exhaustive search.}


All models, including individual and ensemble variants, are selected based on their performance on the validation dataset. {For ensemble models, using the same validation data to tune hyperparameters and identify the optimal combination of models may introduce validation overfitting. Nevertheless, this approach is adopted to ensure that the most recent available information prior to the test period is fully utilized in both hyperparameter optimization and ensemble construction.}

\subsection{Benchmarks and Evaluation}

In order to check the strengths of our models, two state-of-the-art models are selected as benchmarks. The first benchmark is Lasso Estimated AutoRegressive (LEAR), which is a high-dimensional linear benchmark for electricity price forecasting. LEAR coefficients are estimated using L1 regularization (LASSO) to enforce sparsity and prevent overfitting (\cite{lear}). The second benchmark is the Deep Neural Network (DNN), which is a multilayer perceptron with input, output, and two hidden layers. DNN employs ADAM as an optimization algorithm for the weights and {TPE} algorithm to tune the hyperparameters (\cite{dnn}){, where the number of trials are set to 500. }

Both LEAR and DNN are implemented using the Python library epftoolbox (\cite{toolbox}) with a training window size of two years and the following regressors:
\begin{itemize}
\item autoregressive lags of the price ($P_{d-1,h}, P_{d-2,h}, P_{d-3,h}, P_{d-7,h}$),
 \item  day of the week dummies, 
\item renewables as one regressor ($\text{RES}_{i,h} = \text{Solar}_{i,h}+\text{WindOn}_{i,h} +\text{WindOff}_{i,h}$) and $\text{load}_{i,h}$ for $i=d, d-1, d-7$. 
\item {commodity prices ($\text{NGas}_{d-2}$, $\text{Oil}_{d-2}$,
$\text{Coal}_{d-2}$, $\text{EUA}_{d-2}$). Note that those fuel-related prices were not included in the LEAR and DNN discussed in \cite{a}. They are added here to offer a fair comparison between the benchmark and proposed models by employing a similar set of features.}
\end{itemize}

{For DNN, the hyperparameter optimization is performed over the following search space:  the number of hidden neurons, activation function, weight initialization scheme, random seed, dropout rate, learning rate, scaling strategies, optional L1 regularization, and binary selection of regressors.}  

{In contrast, the developed models tune a reduced set of hyperparameters, such that the only common hyperparameters with DNN are the number of neurons, the learning rate, and L1 regularization, which is systematically applied rather than treated as optional.  Other tuned DNN hyperparameters are either predetermined (e.g., activation function) in the proposed models or not part of the modeling framework (e.g., dropout mechanism).  The random seed and weight initialization are also not considered in the developed models, as their influence is limited to the first training window, which is considered minimal. 
}

{During the evaluation stage, both LEAR and DNN are recalibrated daily, and their forecast performance is evaluated using the test data. LEAR does not go through the hyperparameter tuning stage as it has a single hyperparameter, namely the Lasso regularization parameter, which  is recalibrated daily during the evaluation stage. This approach enhances LEAR's predictive accuracy but also increases the training time relative to a standard Lasso model with a fixed penalty determined earlier. DNN hyperparameters, on the other hand, are selected using the validation data, where the network is trained only once and not recalibrated daily.}

{It is worth emphasizing that throughout this study, the benchmark models employ the exact same test datasets as the proposed models.  However, the validation datasets differ: the proposed models use the explicitly specified validation dataset in the beginning of Section \ref{section4.2}, whereas the benchmark model, DNN, use the 42 weeks preceding the test data. Additionally, within this study, one year is defined as 365 days, except when discussing the window size of LEAR and DNN where one year is defined as 364 days to remain consistent with the code provided in the epftoolbox library.}

{The proposed models and DNN differ in architecture, hyperparameter search space, and validation dataset. These differences arise from the specific characteristics and standard implementation requirements of each modeling framework. 
Most importantly, all models are evaluated using exactly the same test datasets. Therefore, the comparison of their out-of-sample forecasting performance remains fair and consistent.}

{However, the comparison of computational time may not be fully consistent, since only the proposed models incorporate the introduced partial online learning mechanism, whereas the benchmark models do not. To address this limitation and strengthen the benchmarking framework, an additional benchmark model employing online learning is included. The new model is the generalized additive model (GAM) equipped with the online-learning algorithm proposed by \textcite{wood2015generalized} and is {referred to as} GAM Online throughout this work. This model retains the upper-triangular factor from the QR decomposition of the model matrix, together with the transformed response quantities needed for estimation, rather than storing the full model matrix. These quantities are built up incrementally from subblocks of the data. Hence, when new data arrive, only the new subblock needs to be processed to update the stored factorization and associated quantities, making re-estimation substantially faster than refitting the model from scratch.}

{For each hour, GAM Online  is fitted using the MGCV package (\cite{wood2015package}) in R, employing the same regressors as those specified in Equation \ref{eq1}. Two hyperparamters are tuned for this model, which are the window size and the number of spline basis functions for each regressor, using the TPE algorithm with 500 trials on the same validation dataset as the proposed models. The model is then evaluated on the same test set as the other models. During both tuning and evaluation stages, GAM Online is re-fitted daily.}

Finally, three accuracy metrics are selected to evaluate and compare the performance of all discussed models: the
root mean square error (RMSE), the mean absolute error (MAE), and the relative mean absolute error (rMAE)
\begin{align*}
\text{RMSE} &= \sqrt{ \frac{1}{24 N_d} \sum_{d=1}^{N_d} \sum_{h=0}^{23} \big( \hat{P}_{d,h}^{\text{model}} - P_{d,h} \big)^2 } \\
\text{MAE}  &= \frac{1}{24 N_d} \sum_{d=1}^{N_d} \sum_{h=0}^{23} \left| \hat{P}_{d,h}^{\text{model}} - P_{d,h} \right| \\
\text{rMAE} &= \frac{
\frac{1}{24 N_d} \sum_{d=1}^{N_d} \sum_{h=0}^{23} \left| \hat{p}_{d,h}^{\text{model}} - p_{d,h} \right|
}{
\frac{1}{24 N_d} \sum_{d=1}^{N_d} \sum_{h=0}^{23} \left| \hat{p}_{d,h}^{\text{naive}} - p_{d,h} \right|
}
\end{align*}

where $\hat{p}_{d,h}^{\text{model}}$ is the forecasted price of the specific model while ${p}_{d,h}$ is the real price. $N_d$ is the number of days in the test period which is set 730 days. $\hat{p}_{d,h}^{\text{naive}}$ is the forecast price of the following naive model:
\[
\hat{p}_{d,h}^{\text{naive}} =
\begin{cases}
p_{d-1,h}, & \text{if $d$ is Tuesday, Wednesday, Thursday, or Friday}, \\
p_{d-7,h}, & \text{if $d$ is Saturday, Sunday, or Monday}.
\end{cases}
\]

For each of the above accuracy metrics, the model with the smallest value of the corresponding metric is considered the best-performing model. However, there is no guarantee that this apparent superiority is statistically significant and could simply result from random chance. Hence, along with the accuracy metrics, the Diebold-Mariano (DM) test is employed, where the DM test is a statistical test that compares the forecast accuracy of two forecasting models (\cite{dm}). Let $\hat{P}_{d,h}^{(A)}$ and $\hat{P}_{d,h}^{(B)}$ be the generated forecast at day $d$ on hour $h$ using models $A$ and $B$, respectively. The forecast error of model $A$ is $\hat{e}_{d}^{(A)}=\left(\sum_{h=0}^{23} \hat{P}_{d,h}^{(A)}-{\sum_{h=0}^{23}{P}_{d,h}}\right)$ and forecast error of model $B$ is $\hat{e}_{d}^{(B)}=\left(\sum_{h=0}^{23} \hat{P}_{d,h}^{(B)}-{\sum_{h=0}^{23}{P}_{d,h}}\right)$.

The null hypothesis to compare the forecast accuracy of $A$ and $B$ is {expressed} as {$E\left(\Delta_d\right)=0,$  where $E(.)$ is the expectation, $\Delta_d=\left|\hat{e}_{d}^{(A)}\right|-\left|\hat{e}_{d}^{(B)}\right|$ under L1 loss and $\Delta_d=\left(\hat{e}_{d}^{(A)}\right)^{2}-\left(\hat{e}_{d}^{(B)}\right)^{2}$ under L2 loss.} The null hypothesis implies that there is no difference in forecast accuracy between $A$ and $B$ {under the assumption that the loss differential, $\Delta_d$, is stationary with short memory.} The alternative hypothesis is expressed as:
{$E\left(\Delta_d\right)<0,$ which suggests that the forecasts of model $A$  are statistically significantly more accurate than those of $B$.} 

{The Diebold–Mariano test is implemented using the small-sample correction proposed by \citet{harvey1997testing} to control for finite-sample effects, and the long-run variance of the test statistic is estimated using the heteroskedasticity- and autocorrelation-consistent approach of \citet{newey1987} with a lag truncation parameter of 7 to account for weekly seasonality.}

\section{Results and Discussion}\label{section5}

\subsection{{Forecasting Accuracy}}
\begin{figure}[h!]
\centering
\includegraphics[width=.95\linewidth]{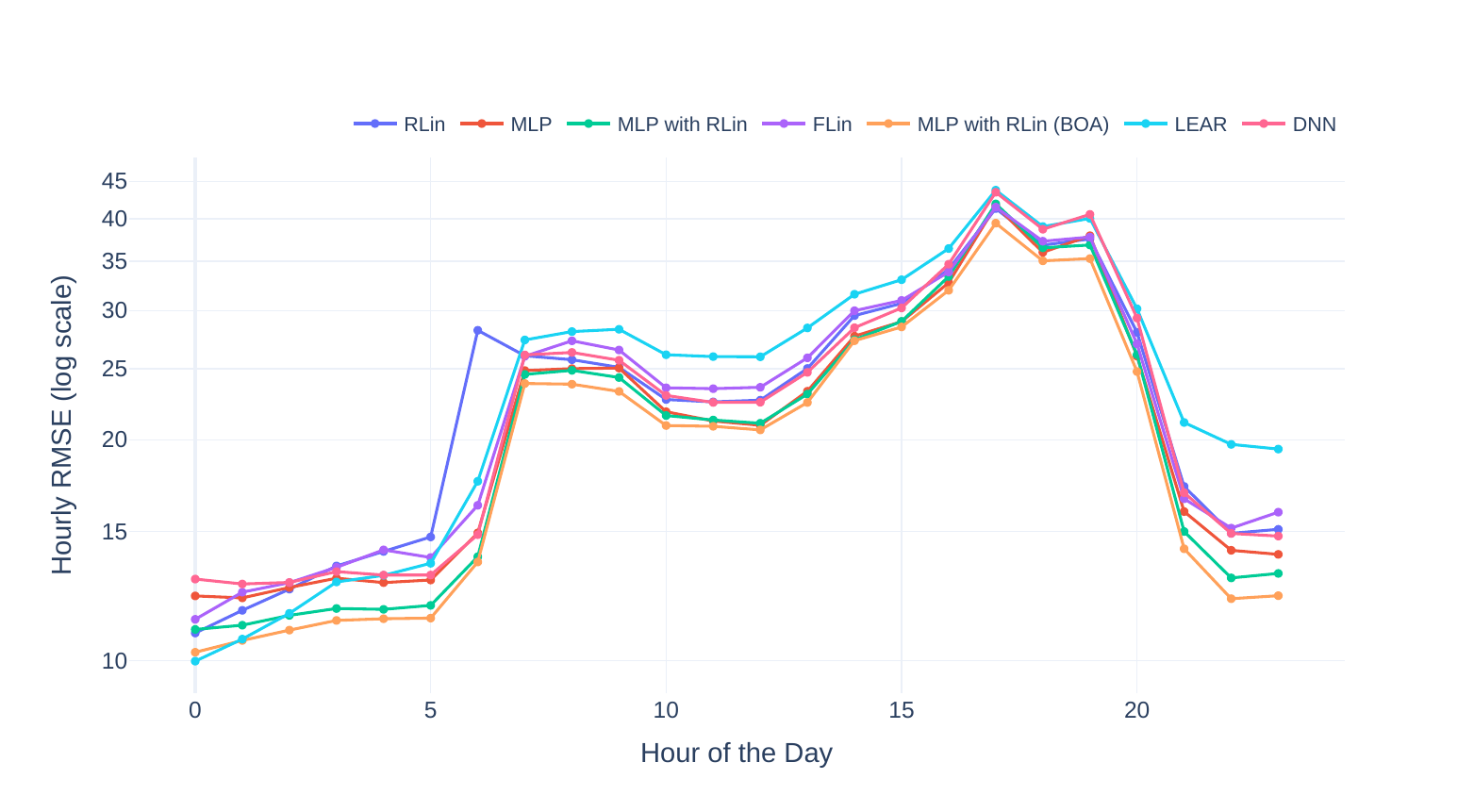}
\caption{Hourly RMSE of selected models for German-Luxembourg bidding zone}
\label{fig3}
\end{figure}

After selecting the optimal models based on their performance on each of the validation datasets, the forecast accuracy of those models is evaluated on the corresponding test data employing accuracy metrics and a statistical test. Figure \ref{fig3} depicts the hourly RMSE of the main models, along with benchmark models for the German-Luxembourg market. Studying Figure \ref{fig3}, one can notice that all models follow the same pattern: perform well during the first and last hours of the day, whereas their performance deteriorates during the peak hours.

During the first hour of the day, the LEAR has the lowest RMSE. However, its performance worsened afterwards. The second benchmark model, which is DNN, performs poorly in most hours of the day. Except for the hour zero,  {\textbf{MLP with RLin (BOA)} generates the most accurate forecast compared to other models as it has the lowest RMSE.}

It is apparent that the model performance changes from one hour to another. Hence, to select the best performing model, three overall accuracy measures are calculated and the results are presented in Table \ref{fig4}.

\begin{table}[h!]
\resizebox{\textwidth}{!}{
\begin{tabular}{lllllll}
\toprule
 & \multicolumn{3}{c}{Germany} & \multicolumn{3}{c}{Spain} \\
 \cmidrule(lr){2-4}\cmidrule(lr){5-7}
 & RMSE & MAE & rMAE & RMSE & MAE & rMAE \\
\midrule
RLin & \cellcolor{rgb,255:red,255;green,250;blue,250}24.94 & \cellcolor{rgb,255:red,255;green,254;blue,254}13.86 & \cellcolor{rgb,255:red,255;green,254;blue,254}38.01\% & \cellcolor{rgb,255:red,255;green,240;blue,240}16.59 & \cellcolor{rgb,255:red,252;green,252;blue,255}11.97 & \cellcolor{rgb,255:red,252;green,252;blue,255}38.65\% \\
MLP & \cellcolor{rgb,255:red,221;green,221;blue,255}23.66 & \cellcolor{rgb,255:red,228;green,228;blue,255}13.07 & \cellcolor{rgb,255:red,228;green,228;blue,255}35.82\% & \cellcolor{rgb,255:red,226;green,226;blue,255}15.76 & \cellcolor{rgb,255:red,227;green,227;blue,255}11.57 & \cellcolor{rgb,255:red,227;green,227;blue,255}37.36\% \\
MLP with RLin & \cellcolor{rgb,255:red,212;green,212;blue,255}23.37 & \cellcolor{rgb,255:red,206;green,206;blue,255}12.39 & \cellcolor{rgb,255:red,206;green,206;blue,255}33.96\% & \cellcolor{rgb,255:red,236;green,236;blue,255}15.94 & \cellcolor{rgb,255:red,235;green,235;blue,255}11.69 & \cellcolor{rgb,255:red,235;green,235;blue,255}37.76\% \\
FLin & \cellcolor{rgb,255:red,255;green,254;blue,254}24.81 & \cellcolor{rgb,255:red,249;green,249;blue,255}13.68 & \cellcolor{rgb,255:red,249;green,249;blue,255}37.50\% & \cellcolor{rgb,255:red,255;green,241;blue,241}16.56 & \cellcolor{rgb,255:red,255;green,245;blue,245}12.16 & \cellcolor{rgb,255:red,255;green,245;blue,245}39.26\% \\
MLP with FLin & \cellcolor{rgb,255:red,205;green,205;blue,255}23.12 & \cellcolor{rgb,255:red,204;green,204;blue,255}12.33 & \cellcolor{rgb,255:red,204;green,204;blue,255}33.79\% & \cellcolor{rgb,255:red,230;green,230;blue,255}15.84 & \cellcolor{rgb,255:red,231;green,231;blue,255}11.63 & \cellcolor{rgb,255:red,231;green,231;blue,255}37.57\% \\
RLin with OLS & \cellcolor{rgb,255:red,252;green,252;blue,255}24.70 & \cellcolor{rgb,255:red,244;green,244;blue,255}13.52 & \cellcolor{rgb,255:red,244;green,244;blue,255}37.06\% & \cellcolor{rgb,255:red,255;green,253;blue,253}16.33 & \cellcolor{rgb,255:red,243;green,243;blue,255}11.83 & \cellcolor{rgb,255:red,243;green,243;blue,255}38.20\% \\
MLP with RLin and OLS & \cellcolor{rgb,255:red,208;green,208;blue,255}23.21 & \cellcolor{rgb,255:red,205;green,205;blue,255}12.36 & \cellcolor{rgb,255:red,205;green,205;blue,255}33.89\% & \cellcolor{rgb,255:red,255;green,216;blue,216}17.04 & \cellcolor{rgb,255:red,255;green,235;blue,235}12.32 & \cellcolor{rgb,255:red,255;green,235;blue,235}39.79\% \\
\hline
RLin (BOA) & \cellcolor{rgb,255:red,219;green,219;blue,255}23.58 & \cellcolor{rgb,255:red,215;green,215;blue,255}12.65 & \cellcolor{rgb,255:red,215;green,215;blue,255}34.69\% & \cellcolor{rgb,255:red,230;green,230;blue,255}15.83 & \cellcolor{rgb,255:red,211;green,211;blue,255}11.32 & \cellcolor{rgb,255:red,211;green,211;blue,255}36.55\% \\
MLP (BOA) & \cellcolor{rgb,255:red,178;green,178;blue,255}\textbf{22.21} & \cellcolor{rgb,255:red,182;green,182;blue,255}11.66 & \cellcolor{rgb,255:red,182;green,182;blue,255}31.98\% & \cellcolor{rgb,255:red,185;green,185;blue,255}14.97 & \cellcolor{rgb,255:red,186;green,186;blue,255}10.92 & \cellcolor{rgb,255:red,186;green,186;blue,255}35.28\% \\
MLP with RLin (BOA) & \cellcolor{rgb,255:red,186;green,186;blue,255}22.48 & \cellcolor{rgb,255:red,182;green,182;blue,255}11.66 & \cellcolor{rgb,255:red,182;green,182;blue,255}31.98\% & \cellcolor{rgb,255:red,178;green,178;blue,255}\textbf{14.84} & \cellcolor{rgb,255:red,178;green,178;blue,255}\textbf{10.79} & \cellcolor{rgb,255:red,178;green,178;blue,255}\textbf{34.84\%} \\
FLin (BOA) & \cellcolor{rgb,255:red,212;green,212;blue,255}23.37 & \cellcolor{rgb,255:red,214;green,214;blue,255}12.65 & \cellcolor{rgb,255:red,214;green,214;blue,255}34.67\% & \cellcolor{rgb,255:red,213;green,213;blue,255}15.50 & \cellcolor{rgb,255:red,208;green,208;blue,255}11.27 & \cellcolor{rgb,255:red,208;green,208;blue,255}36.41\% \\
MLP with FLin (BOA) & \cellcolor{rgb,255:red,205;green,205;blue,255}23.12 & \cellcolor{rgb,255:red,188;green,188;blue,255}11.84 & \cellcolor{rgb,255:red,188;green,188;blue,255}32.47\% & \cellcolor{rgb,255:red,178;green,178;blue,255}14.84 & \cellcolor{rgb,255:red,178;green,178;blue,255}10.79 & \cellcolor{rgb,255:red,178;green,178;blue,255}34.84\% \\
RLin with OLS (BOA) & \cellcolor{rgb,255:red,217;green,217;blue,255}23.52 & \cellcolor{rgb,255:red,212;green,212;blue,255}12.58 & \cellcolor{rgb,255:red,212;green,212;blue,255}34.49\% & \cellcolor{rgb,255:red,222;green,222;blue,255}15.67 & \cellcolor{rgb,255:red,207;green,207;blue,255}11.25 & \cellcolor{rgb,255:red,207;green,207;blue,255}36.35\% \\
MLP with RLin and OLS (BOA) & \cellcolor{rgb,255:red,187;green,187;blue,255}22.51 & \cellcolor{rgb,255:red,178;green,178;blue,255}\textbf{11.55} & \cellcolor{rgb,255:red,178;green,178;blue,255}\textbf{31.65\%} & \cellcolor{rgb,255:red,182;green,182;blue,255}14.91 & \cellcolor{rgb,255:red,180;green,180;blue,255}10.82 & \cellcolor{rgb,255:red,180;green,180;blue,255}34.96\% \\
BOA all & \cellcolor{rgb,255:red,201;green,201;blue,255}22.97 & \cellcolor{rgb,255:red,187;green,187;blue,255}11.83 & \cellcolor{rgb,255:red,187;green,187;blue,255}32.42\% & \cellcolor{rgb,255:red,183;green,183;blue,255}14.94 & \cellcolor{rgb,255:red,180;green,180;blue,255}10.82 & \cellcolor{rgb,255:red,180;green,180;blue,255}34.95\% \\
\hline
LEAR & \cellcolor{rgb,255:red,255;green,201;blue,201}26.57 & \cellcolor{rgb,255:red,255;green,178;blue,178}16.16 & \cellcolor{rgb,255:red,255;green,178;blue,178}44.30\% & \cellcolor{rgb,255:red,255;green,178;blue,178}17.77 & \cellcolor{rgb,255:red,255;green,178;blue,178}13.24 & \cellcolor{rgb,255:red,255;green,178;blue,178}42.75\% \\
DNN & \cellcolor{rgb,255:red,255;green,249;blue,249}24.97 & \cellcolor{rgb,255:red,255;green,251;blue,251}13.95 & \cellcolor{rgb,255:red,255;green,251;blue,251}38.25\% & \cellcolor{rgb,255:red,255;green,222;blue,222}16.93 & \cellcolor{rgb,255:red,255;green,221;blue,221}12.54 & \cellcolor{rgb,255:red,255;green,221;blue,221}40.51\% \\
GAM Online & \cellcolor{rgb,255:red,255;green,178;blue,178}27.35 & \cellcolor{rgb,255:red,255;green,205;blue,205}15.34 & \cellcolor{rgb,255:red,255;green,205;blue,205}42.05\% & \cellcolor{rgb,255:red,255;green,207;blue,207}17.22 & \cellcolor{rgb,255:red,255;green,185;blue,185}13.13 & \cellcolor{rgb,255:red,255;green,185;blue,185}42.40\% \\
\bottomrule
\end{tabular}
}
\caption{Comparative forecast accuracy in terms of RMSE (EUR/MWh), MAE (EUR/MWh) and rMAE of competing Models. For each metric, the cell colors range from red (indicating poorer performance) to blue (indicating better performance). The best performing models are highlighted in bold.}
\label{fig4}
\end{table}

{For both the German-Luxembourg and Spanish bidding zones, ensemble models generally outperform individual and benchmark models. For instance, in the Spanish market, \textbf{MLP with RLin (BOA)} achieves the lowest RMSE, MAE, and rMAE among all evaluated models.  Overall, most of the models developed in this paper outperform the benchmark models in both markets. In the German market, even a linear model such as \textbf{FLin} demonstrates better forecasting performance than the nonlinear benchmark DNN.}

{The RMSE is reduced by 11-12\% and the MAE by 14-17\% when comparing the best model to DNN in the respective market. RMSE and MAE are also reduced when comparing BOA models to their individual counterparts. For instance,  there is a  4\%  to  11\% reduction in MAE in the German Luxembourg market. The gains from ensembling vary across model classes and may be attributed to differences in the correlation between forecast errors, as models with different structures usually benefit more from ensembling. For example, the empirical results of \textcite{marcjasz2018selection}  and \textcite{hubicka2018note}  suggest that combining models trained on short and long training windows improves accuracy.}

Additionally, one can notice an overall decrease in both RMSE and MAE in the Spanish market compared to the same models in the German-Luxembourg market. This might be the result of the share of renewables in the electricity generation. \textcite{rintamaki2017does} confirmed that wind-powered generation increases the price volatility in Germany. On the other hand, \textcite{pereira2017effect} concluded that while wind increases volatility in Spain, the integration of hydro power plants decreases it. 

\begin{figure}[h!]
\centering
\includegraphics[width=1.0\linewidth]{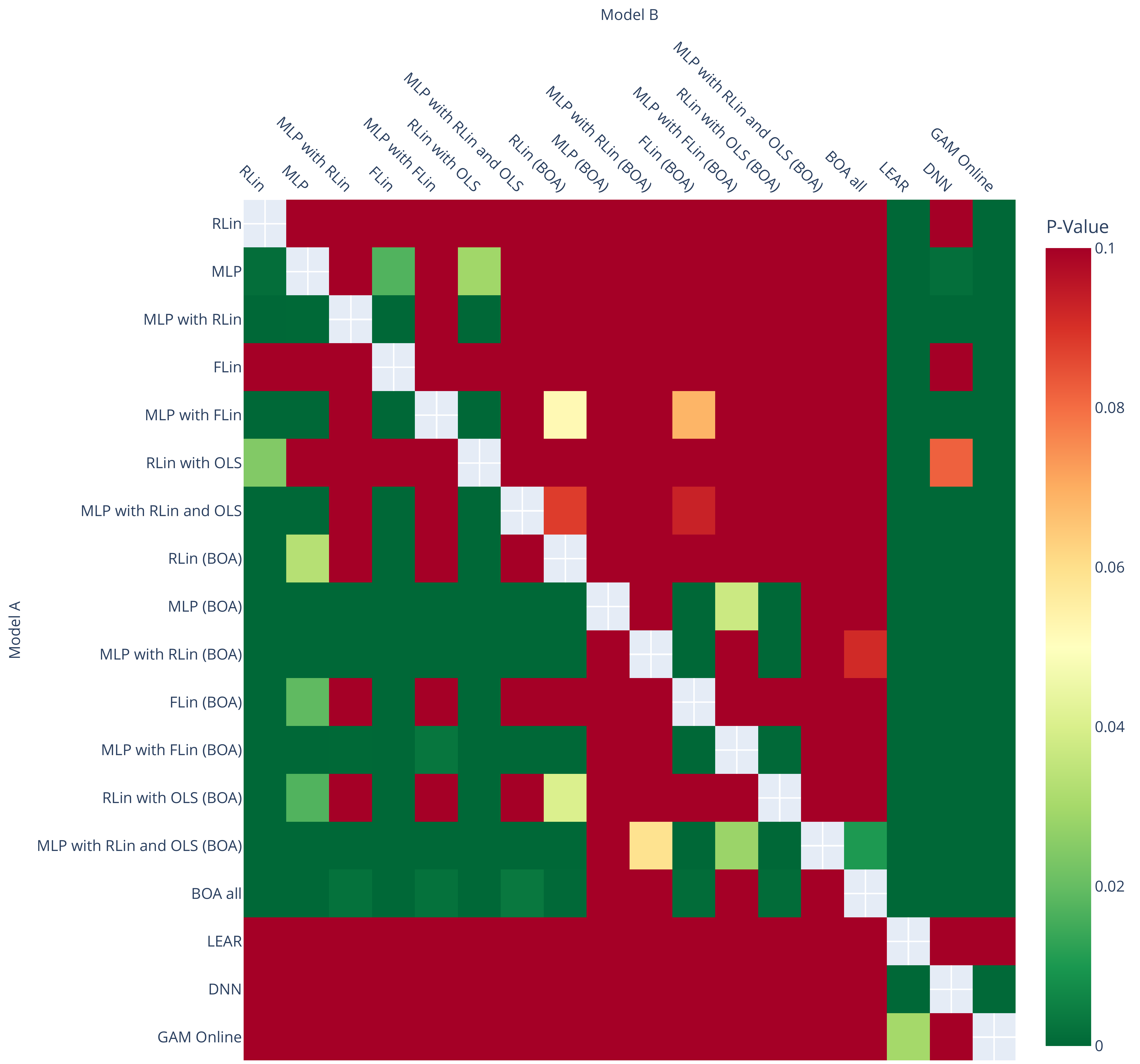}
\caption{Pairwise DM test p-values across forecasting models for German-Luxembourg market {using L1 loss}}
\label{fig5}
\end{figure}

Figure  \ref{fig5} presents the heatmap of the p-value of the Diebold Mariano test for the German-Luxembourg bidding zone { (see Figure \ref{fig16} in the Appendix for Spain) under the absolute loss.} The horizontal model represents model A, while the vertical one represents model B. With a 5\% significance level, the null hypothesis is rejected if the p-value is less than 0.05. Based on the legend, shades of green indicate rejection of the null, whereas shades of red indicate failure to reject. The stationarity assumption underlying the DM test was verified using the Augmented Dickey–Fuller (ADF) test (\cite{dickey1979distribution}). The ADF test results, presented in Figures \ref{sta1}, \ref{sta2}, \ref{sta3}, and \ref{sta4} in the Appendix, confirm that the stationarity assumption is satisfied. {These results were obtained using a specification with an intercept and a maximum lag length of seven, with the lag order selected using the Akaike information criterion (AIC). Alternative specifications, including no constant or trend, a constant with a linear trend, and a constant with linear and quadratic trends, led to the same conclusion of stationarity}

{Based on Figure \ref{fig5}, } \textbf{MLP with RLin and OLS (BOA)} is more accurate than any other model except \textbf{MLP (BOA)} and \textbf{MLP with RLin (BOA)}, with the difference being statistically significant. On the other hand, the benchmark models perform poorly. The green vertical cells for benchmarks in Figure \ref{fig5} imply that most of the proposed models generates statistically significantly more accurate forecasts than the benchmark models, confirming the previous observation that these models are the worst-performing ones. Moreover, the horizontal cells for the benchmark models indicate that DNN is only statistically significantly more precise than LEAR {and GAM Online}, whereas LEAR is not statistically significantly more accurate than any other model.

{Since the loss function is a key component of the DM test, we implement the DM test under squared loss to examine whether the conclusions about predictive ability change when changing losses. The DM output under the squared loss for different markets are dipicted in Figure \ref{fig15} and \ref{fig17} in the Appendix, where the majority of the developed models exhibit statistically significant improvements in forecast accuracy compared to the benchmark models. This confirms that the superior forecasting performance of our models relative to the benchmarks is robust across different markets and choices of loss functions.}

{It is important to emphasize that the reported DM test results are based on daily aggregated forecast errors rather than hourly errors. Applying the DM test separately at the hourly level may reveal that different models perform best at different hours, which would make it infeasible to identify a single best-performing model across all hours. Additionally, conducting numerous DM tests simultaneously may increase the risk of false-positive findings due to multiple testing. However, no correction for multiple testing is applied, as this is not common practice in the electricity price forecasting literature.
}

\subsection{{Computational Cost vs Accuracy}}

\begin{figure}[h!]
\centering
\includegraphics[width=1.0\linewidth]{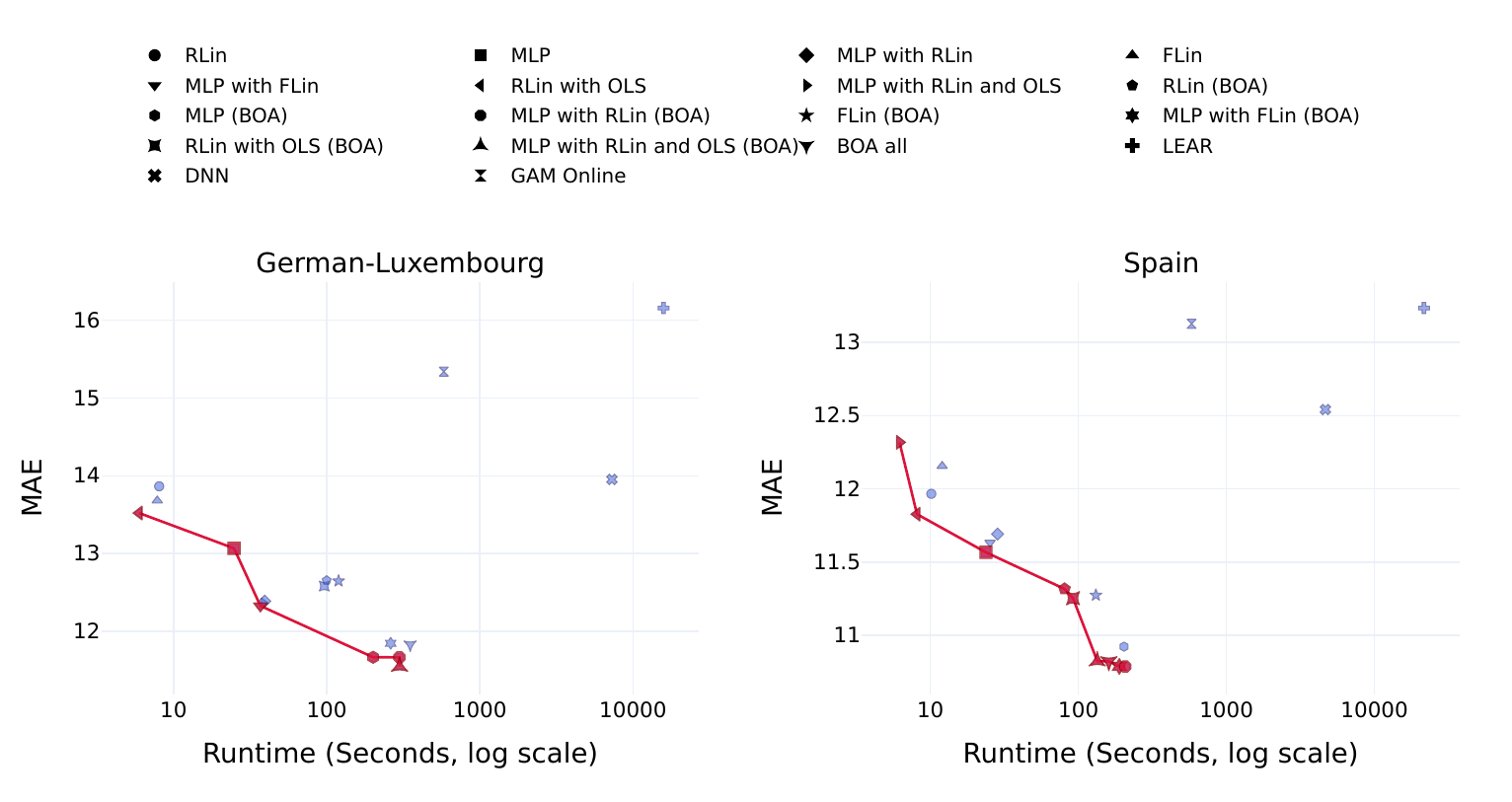}
\caption{
{Runtime and MAE of the considered forecasting models with the efficient Pareto frontier. The red markers correspond to Pareto-optimal models, while the blue markers represent non-Pareto models.}}
\label{fig6}
\end{figure}

Figure  \ref{fig6} illustrates the trade-off between the runtime of a model and its accuracy, {where runtime denotes the time required to execute the full forecasting study on the test dataset}\footnote{{ The recorded runtime of the proposed and benchmark models was measured on a MacBook Pro (2020) equipped with an Apple M1 processor (8 CPU cores) and 16 GB of unified memory.}}. {The execution time of the proposed models ranges from {6 seconds to about 6 minutes} in the German–Luxembourg market and from {6 seconds to 3 minutes and half} in the Spanish market. The best performing models in terms of time are generally linear models. In contrast, the ensemble models are the slowest with the highest forecast accuracy.} 

While the benchmark models are the worst-performing models in terms of accuracy, their performance in terms of time is much worse. {It took LEAR about one and a half to two hours to train and generate forecasts for the two-year test period, whereas the DNN needed about four and a half hours in the German–Luxembourg market and about six hours in Spain.} {These results are expected, as those models do not incorporate online learning. By contrast, the benchmark model that employs online learning achieves exceptionally short runtimes. For each market, GAM Online requires only about {nine and a half minutes} to train and generate forecasts. However, this computational advantage comes at the cost of lower forecasting accuracy. On the other hand, despite having a relatively short runtime among the benchmark models, GAM Online remains computationally more expensive than all the proposed models while producing less accurate forecasts.}

\subsection{{Impact of Study Design and Hyperparameters}}

{This section examines the impact of the online approach implementation and evaluates the model’s robustness to variations in hyperparameters and time periods. All the analyses are conducted using one of the best-performing specifications, namely \textbf{MLP with RLin} model. In the time-period analysis, we additionally consider the \textbf{MLP with RLin (BOA)}. All experiments are carried out using the German–Luxembourg dataset.}

{To isolate the effect of the proposed online learning strategy on forecasting accuracy and, in particular, its contribution to reducing computational cost, we train the model under three different settings without online learning. Specifically, no warm start is applied and a fixed set of hyperparameters is used for all training windows (initial window and updated windows). The forecasting accuracy and computational time obtained without online learning are then compared with those achieved using online learning.}

{In the first setting, \textbf{MLP with RLin} is trained using the hyperparameters of the initial window of the online learning, i.e., the number of epochs set to 60, and the remaining hyperparameters are tuned based on the range specified in the first column of Table \ref{tab:hyperparameters} in the Appendix. This model is referred to as \textbf{MLP with RLin (Initial)}. The second setting is where the model is trained using the hyperparameter of updated windows, that is, the number of epochs is set to 10 and the other hyperparameters are tuned according to the Range$_{up}$ column in Table \ref{tab:hyperparameters}. This model is {referred to} as \textbf{MLP with RLin (Update)}. The last setting is similar to the first one, with the difference that the number of epochs is treated as a tunable hyperparameter, with a range from 5 to 75. This model is named \textbf{MLP with RLin (Tuned)}}.

{The above three discussed models without online learning are compared with the original model that employs online learning, hereafter {referred to} as \textbf{MLP with RLin}. The corresponding results are reported in Table \ref{No_online}.}

\begin{table}[h!]
\centering
\begin{tabular}{lccc}
\toprule
Model & MAE & RMSE & Runtime (sec) \\
\midrule
MLP with RLin & \cellcolor{rgb,255:red,178;green,178;blue,255}\textbf{12.39} & \cellcolor{rgb,255:red,178;green,178;blue,255}\textbf{23.37} & \cellcolor{rgb,255:red,192;green,192;blue,255}39.41 \\
MLP with RLin (Tuned) & \cellcolor{rgb,255:red,214;green,214;blue,255}12.78 & \cellcolor{rgb,255:red,194;green,194;blue,255}23.72 & \cellcolor{rgb,255:red,255;green,178;blue,178}278.07 \\
MLP with RLin (Update) & \cellcolor{rgb,255:red,217;green,217;blue,255}12.81 & \cellcolor{rgb,255:red,202;green,202;blue,255}23.90 & \cellcolor{rgb,255:red,255;green,221;blue,221}203.74 \\
MLP with RLin (Initial) & \cellcolor{rgb,255:red,255;green,178;blue,178}14.05 & \cellcolor{rgb,255:red,255;green,178;blue,178}26.69 & \cellcolor{rgb,255:red,178;green,178;blue,255}\textbf{16.18} \\
\bottomrule
\end{tabular}
\caption{{Performance comparison with and without online learning. Runtime represents the time (in seconds) required to complete the full forecasting study across the 730 training windows in the test dataset. For each column the cell colors range from red (indicating poorer performance) to blue (indicating better performance). The best performing models are highlighted in bold.}}
\label{No_online}
\end{table}


\textbf{MLP with RLin (Initial)} is the best-performing model in terms of computational time, while it has the worst performance in terms of accuracy. \textbf{MLP with RLin (update)} and \textbf{MLP with RLin (tuned)} are much better in terms of forecast performance, but they require a long time to train the model. {However, the last two models remain faster than all benchmark models, including the benchmark with online learning, while also producing more accurate forecasts. This confirms the effectiveness of the employed architecture in producing efficient forecasts.}

{On the other hand, \textbf{MLP with RLin}, which incorporates the proposed online learning approach, is approximately five times as fast as the non-online variants, except for \textbf{MLP with RLin (Initial)}, while also achieving higher forecasting accuracy. Those results highlight the importance of the proposed partial online learning strategy in reducing computational cost and improving forecast accuracy. This outcome is particularly noteworthy because \textcite{ash2020warm} argues that, although warm-start strategies reduce computational time, they may adversely affect generalization performance. In contrast, our empirical results show that combining warm-starting with window-specific adaptation yields simultaneous improvements in forecasting accuracy and computational efficiency.}

\begin{figure}[h!]
\centering
\includegraphics[width=1.0\linewidth]{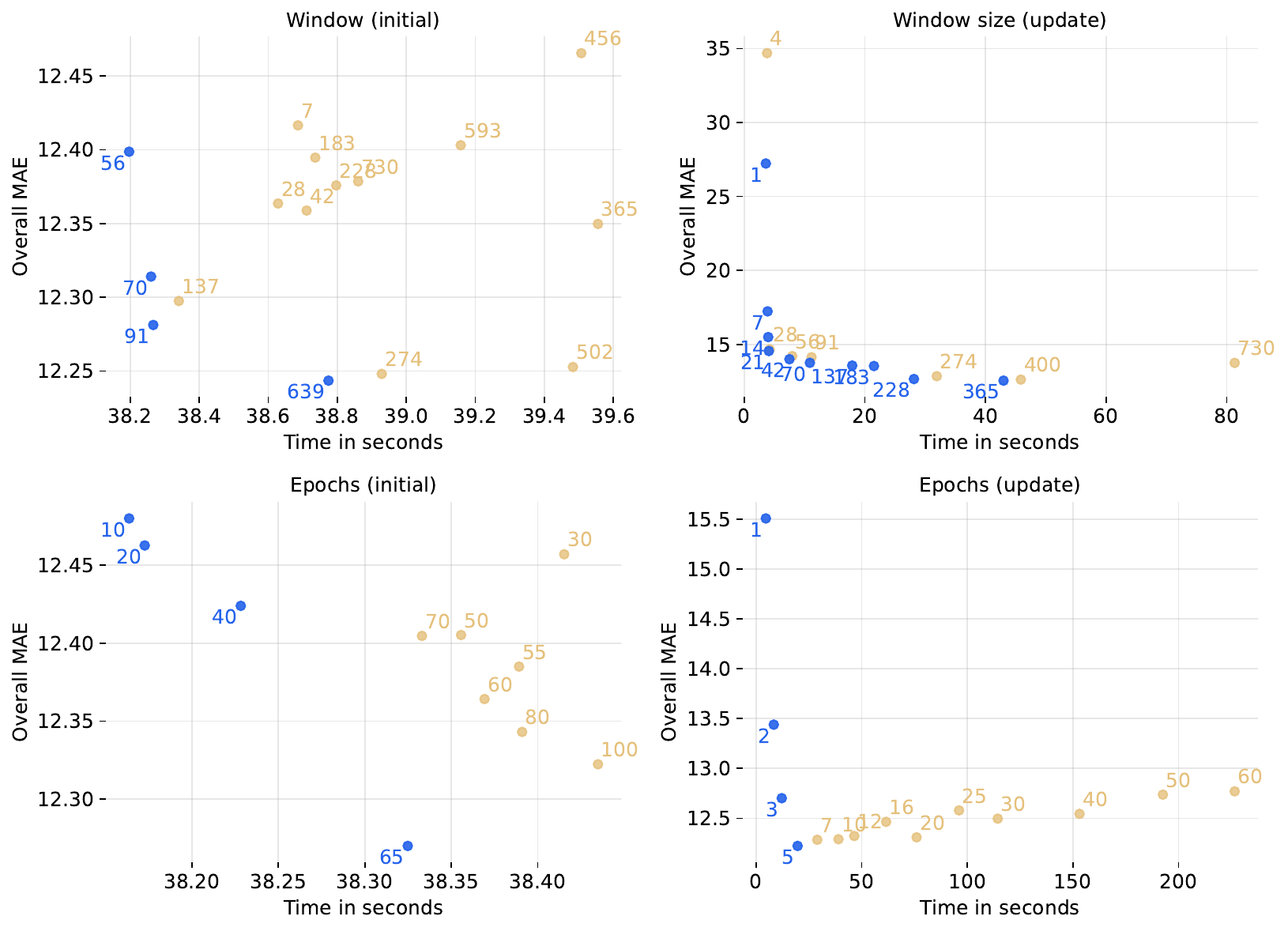}
\caption{{Sensitivity analysis of the main hyperparameters. Blue markers indicate Pareto-optimal configurations, whereas gold markers denote dominated ones.}
}
\label{hyper_para}
\end{figure}

{To investigate the impact of hyperparameters on model performance, the following analysis is conducted. The optimal hyperparameters selected based on the validation dataset are fixed, while the hyperparameter under study is allowed to vary. Each time the model is trained with a different value of the free hyperparameter from a predefined grid to examine how forecasting accuracy and computational time change across different configurations. Figure \ref{hyper_para} displays the results of the sensitivity analysis for the four main hyperparameters, examined separately.}

{ The left two panels show the results for window size and number of epochs for the initial training window.  These two hyperparameters seem to have a limited impact on performance, as the time and accuracy do not vary significantly when the hyperparameters change across the considered values. On the other hand, the window size and number of epochs for the updated training window (right panels) affect the model performance substantially. For instance, when the window size is 1 (corresponding to the case of pure online learning), the MAE is 27 with a running time of less than 4 seconds only. 
Later, one can see a clear trend where the increase in window size is associated with a decrease in MAE and an increase in computation time. A similar trend can be observed in the number of epochs for the updated training window.}

{After investigating the effect of online learning and hyperparameters on the performance of the model, we now assess the robustness of the model across different time periods. Throughout this work, the analysis of all model are carried out using a validation and test dataset specified in Subsection \ref{section4.2}. In the present analysis, however, we modify the data split: the validation dataset covers the period from 2020-01-17 to 2021-01-15, while the test dataset spans the subsequent three years, from 2021-01-16 to 2024-01-15. Train data, on the other hand, are the observations that precede the target observation to be forecasted, following the online scheme framework adopted throughout the study. The importance of the new split is that the test dataset covers one of the largest European energy crises in recent years. This setup, therefore, provides a rigorous evaluation of the model’s robustness under extreme market conditions.}

Using the new validation dataset, the optimal hyperparameters are selected based on the same ranges specified in Table \ref{tab:hyperparameters} in the Appendix, with the exception of $D_{\text{init}}$, whose range is adjusted to 30–365. This modification is necessary due to the reduced amount of training data available under the revised data split. For the same reason, the training-window size for the DNN and LEAR models is reduced from two years to one year{, while the hyperparameter-tuning range for GAM Online is limited to a maximum of one year.
}

{For the new test dataset, which spans a three-year period, RMSE and MAE are evaluated separately for the pre-crisis, crisis (spanning the period from 2021-09-01 to 2023-01-15), and post-crisis subperiods, as presented in Table \ref{crisis}.}

\begin{table}[!ht]
\centering
\resizebox{\textwidth}{!}{
\begin{tabular}{lcccccc}
\toprule
 & \multicolumn{3}{c}{MAE} & \multicolumn{3}{c}{RMSE} \\
\cmidrule(lr){2-4} \cmidrule(lr){5-7}
Model
& Pre-crisis & Crisis & Post-crisis
& Pre-crisis & Crisis & Post-crisis \\
\midrule
MLP with RLin
& \cellcolor{rgb,255:red,198;green,198;blue,255}6.19 & \cellcolor{rgb,255:red,187;green,187;blue,255}24.81 & \cellcolor{rgb,255:red,197;green,197;blue,255}12.27
& \cellcolor{rgb,255:red,204;green,204;blue,255}9.43 & \cellcolor{rgb,255:red,188;green,188;blue,255}34.52 & \cellcolor{rgb,255:red,192;green,192;blue,255}18.08 \\

MLP with RLin (BOA)
& \cellcolor{rgb,255:red,178;green,178;blue,255}\textbf{5.94} & \cellcolor{rgb,255:red,178;green,178;blue,255}\textbf{24.31} & \cellcolor{rgb,255:red,178;green,178;blue,255}\textbf{11.51}
& \cellcolor{rgb,255:red,178;green,178;blue,255}\textbf{8.95} & \cellcolor{rgb,255:red,178;green,178;blue,255}\textbf{33.62} & \cellcolor{rgb,255:red,178;green,178;blue,255}\textbf{17.37} \\

LEAR
& \cellcolor{rgb,255:red,255;green,178;blue,178}7.88 & \cellcolor{rgb,255:red,255;green,188;blue,188}32.54 & \cellcolor{rgb,255:red,255;green,196;blue,196}17.01
& \cellcolor{rgb,255:red,255;green,178;blue,178}11.78 & \cellcolor{rgb,255:red,255;green,201;blue,201}44.73 & \cellcolor{rgb,255:red,255;green,179;blue,179}25.08 \\

DNN
& \cellcolor{rgb,255:red,227;green,227;blue,255}6.56 & \cellcolor{rgb,255:red,255;green,236;blue,236}29.76 & \cellcolor{rgb,255:red,255;green,178;blue,178}17.73
& \cellcolor{rgb,255:red,242;green,242;blue,255}10.13 & \cellcolor{rgb,255:red,255;green,212;blue,212}43.81 & \cellcolor{rgb,255:red,255;green,178;blue,178}25.14 \\

GAM Online
& \cellcolor{rgb,255:red,255;green,231;blue,231}7.21 & \cellcolor{rgb,255:red,255;green,178;blue,178}33.11 & \cellcolor{rgb,255:red,255;green,212;blue,212}16.34
& \cellcolor{rgb,255:red,255;green,242;blue,242}10.59 & \cellcolor{rgb,255:red,255;green,178;blue,178}46.68 & \cellcolor{rgb,255:red,255;green,219;blue,219}23.05 \\
\bottomrule
\end{tabular}
}
\caption{{ MAE and RMSE for the three evaluation periods in the German–Luxembourg market. Pre-crisis period spans 2021-01-16 to 2021-08-31, Crisis period covers 2021-09-01 to 2023-01-15, and Post-crisis period corresponds to 2023-01-16 to 2024-01-15. For each metric and period, the cell colors range from red (indicating poorer performance) to blue (indicating better performance). The best performing models are highlighted in bold.}}
\label{crisis}
\end{table}

{The main finding reported in Table \ref{crisis} is that, although the forecasting accuracy of the proposed models deteriorates during the crisis period, their performance remains substantially superior to that of the benchmark models, with approximately a {18}\% reduction in MAE and {23}\% reduction in RMSE. 
}

{Finally, to address the remaining item in the checklist proposed by \textcite{a} for ensuring adequate electricity price forecasting research, we conduct an additional forecasting study using the Pennsylvania–New Jersey–Maryland (PJM) open dataset available in \textcite{toolbox}, with the period from 27-12-2016 to 24-12-2018 as test data.} 

{The results in Table \ref{tab_PJM} in the Appendix show that the proposed model outperforms DNN in terms of both forecast accuracy and computational efficiency in the PJM market. These findings, together with the preceding results, provide evidence that the effectiveness of the proposed approach is not confined to a specific time period or solely to European markets and may generalize across different time periods and international electricity markets.}

\section{Conclusion}\label{section6}

We {employ a multivariate} approach that combines linear and nonlinear mechanisms to develop a forecasting model for electricity prices with high accuracy and computational efficiency. We implemented a main architecture that has two main components: a skip connection from input to output and a nonlinear part connecting the input to the output through a hidden layer. {Note that the coefficients of the linear skip connection are not directly interpretable because the linear and nonlinear components are not strictly identifiable. However, orthogonalization can be applied to regain interpretability.} 

We adopt an ensemble method, following the fully adaptive BOA approach, to aggregate individual forecast models and produce more accurate predictions. {More importantly}, we propose a novel partial online learning method that substantially reduces computational cost by combining warm starting with window-specific hyperparameters.

{The models’ performance is evaluated against two state-of-the-art benchmarks in day-ahead electricity price forecasting, namely LEAR and DNN, as well as an additional benchmark based on a generalized additive model with online learning, {referred to} as GAM Online. All the analyses are conducted using data from the} German-Luxembourg and Spanish bidding zones, which are among the largest European markets. The measures of accuracy demonstrate the superiority of all the proposed models over the benchmark models with a {11\%}  reduction in RMSE and {17\%} decrease in MAE for the German-Luxembourg market and a decline of {12\%} in RMSE and {14\%} in MAE for Spain. The results of the Diebold Mariano test are consistent with measures of accuracy metrics. All benchmarks fail to outperform any of the proposed models. Moreover, the superiority of the proposed models over the benchmarks remains evident even during the most recent European energy crisis.

In addition to their strong predictive accuracy, the suggested models achieve an outstanding runtime. It took the fast models only a few seconds to fit and generate the forecast, while even the slowest and most accurate models complete training and predictions within a few minutes.  {By contrast, the benchmark models exhibit substantially longer runtimes. Although GAM Online is considerably faster than the other benchmark models, it remains slower than the proposed models.} {Importantly, these runtime results reflect the computational performance of the implemented forecasting pipelines, in which neither LEAR nor DNN is equipped with an online learning mechanism. On the other hand, the contribution of partial online learning to computational efficiency is more directly demonstrated by our ablation analysis, in which we compare the proposed model with and without the partial online learning mechanism. This comparison shows a substantial reduction in runtime when partial online learning is employed.}

In future work, we aim to extend those models to generate probabilistic forecasts due to the limitations of point forecasts in a world full of uncertainties, especially with the large integration of renewables in the grid. Additionally, we plan to explore more advanced neural network architectures such as RNN and LSTM, while carefully monitoring the computational time.

\newpage

\section*{Acknowledgements}
This research was partially funded in the course of TRR 391 Spatio-temporal Statistics for the Transition of Energy and Transport (520388526) by the Deutsche Forschungsgemeinschaft (DFG, German Research Foundation)

\section*{Declaration on AI assisted tools}

During the preparation of this work, the authors used ChatGPT (OpenAI) to improve the readability and clarity of the manuscript and enhance code quality. After using this tool, the authors reviewed and edited the content as needed and take full responsibility for the content of the published article.


\newpage
\addcontentsline{toc}{section}{Bibliography}
\renewcommand\refname{Bibliography} 
\printbibliography
\newpage
\appendix
\addsec{Appendix}
\begin{figure}[htb!]
\centering
\includegraphics[width=0.99\textwidth]{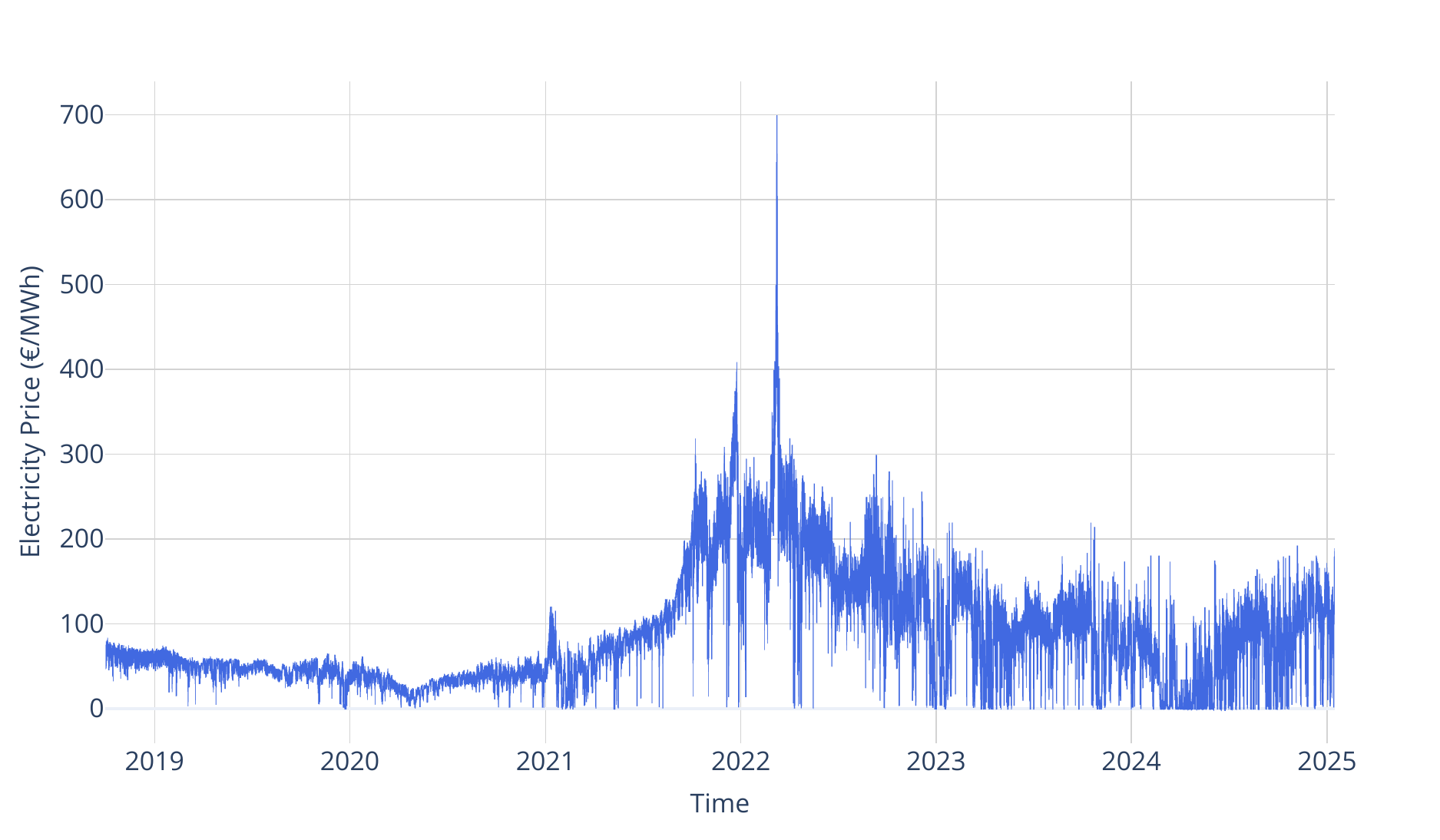}
\caption{Time series of day-ahead electricity prices in Spain}
\end{figure}

\begin{figure}[htb!]
\centering
\includegraphics[width=0.99\textwidth]{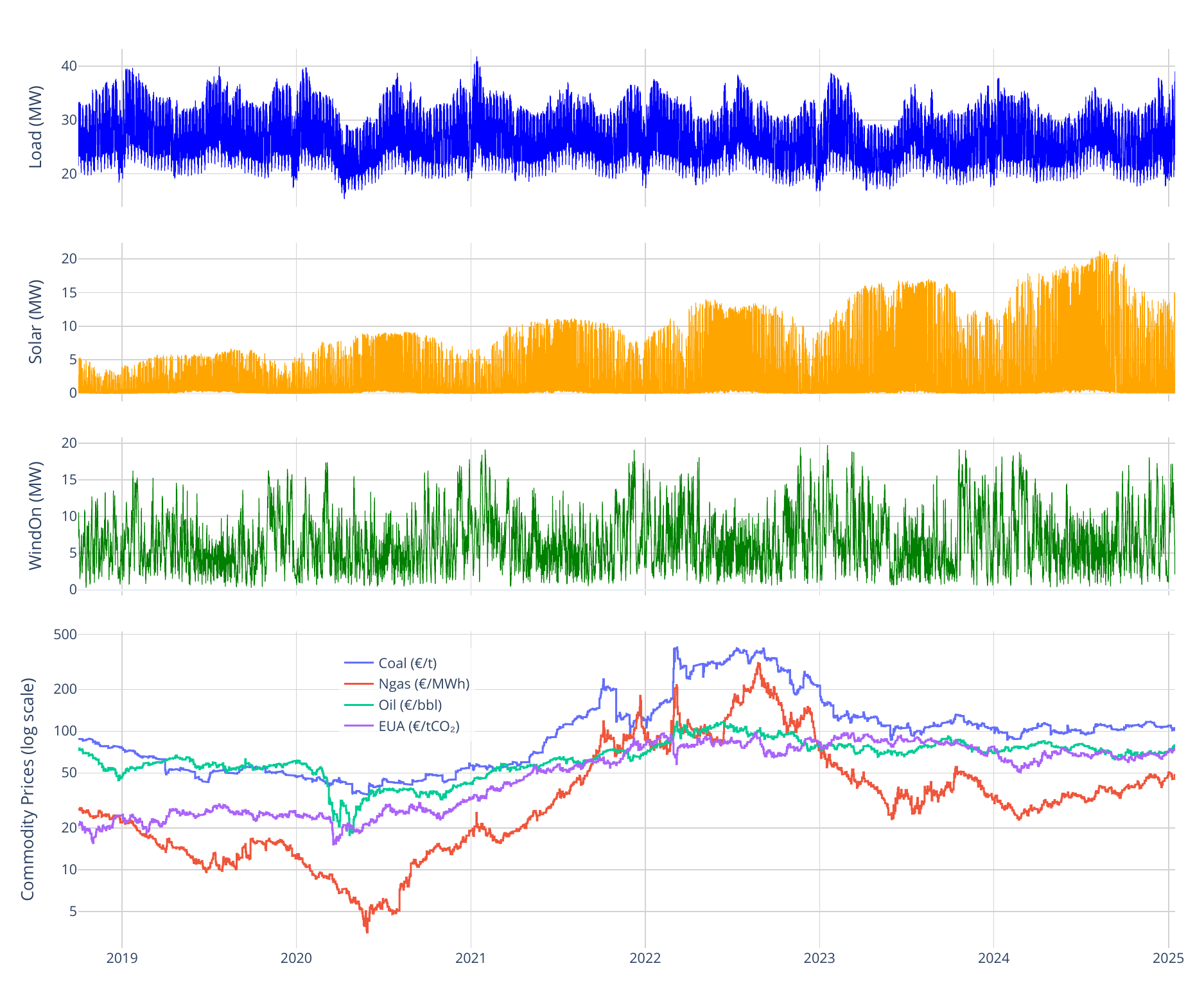}
\caption{Day-ahead forecasts of load and renewable generation, and commodity prices in the Spanish electricity market}
\end{figure}

\begin{table}[ht]
\centering

\begin{tabular}{llllc}
\hline
\textbf{} & \textbf{Range$_{init}$} &\textbf{Range$_{up}$} & \textbf{Applies to} \\
\hline
${lr}$ 
& $[10^{-5},\,10^{-2}]$ 
& $[10^{-4},\,10^{-2}]$
& All models \\

$D$ 
& $[30,\,730]$ 
& $[1,\,365]$  
& All models \\

$\lambda_{\texttt{1}}$ 
& $[10^{-5},\,10^{-2}]$ 
& $[10^{-4},\,10^{-2}]$ 
& All models \\

$\lambda_{\texttt{2}}$ 
& $[10^{-5},\,10^{-2}]$ 
& $[10^{-4},\,10^{-2}]$ 
& All models \\

$\alpha$ 
& $[0,\,2]$ 
& $[0,\,2]$ 
& Only OLS-based models \\

$nr$ 
& $[1,\,128]$ 
& $[1,\,128]$ 
& Only models with MLP \\
\hline
\end{tabular}

\caption{{Hyperparameter search space used in the Optuna optimization. 
The parameters $l r$, $D$, $\lambda_{1}$, and $\lambda_{2}$ denote, respectively, the learning rate, the length of the training window (in days), and the penalty terms in the loss function.
The parameter $\alpha$ controls the contribution of the OLS-based initialization and is only used for models with OLS initialization.
Finally, $nr$ denotes the number of neurons in the hidden layer. Range$_{init}$ corresponds to the tuning range of the initial window, while Range$_{up}$ represents the tuning range of the updated windows.}}

\label{tab:hyperparameters}
\end{table}

\begin{figure}[h!]
\centering
\includegraphics[width=.95\linewidth]{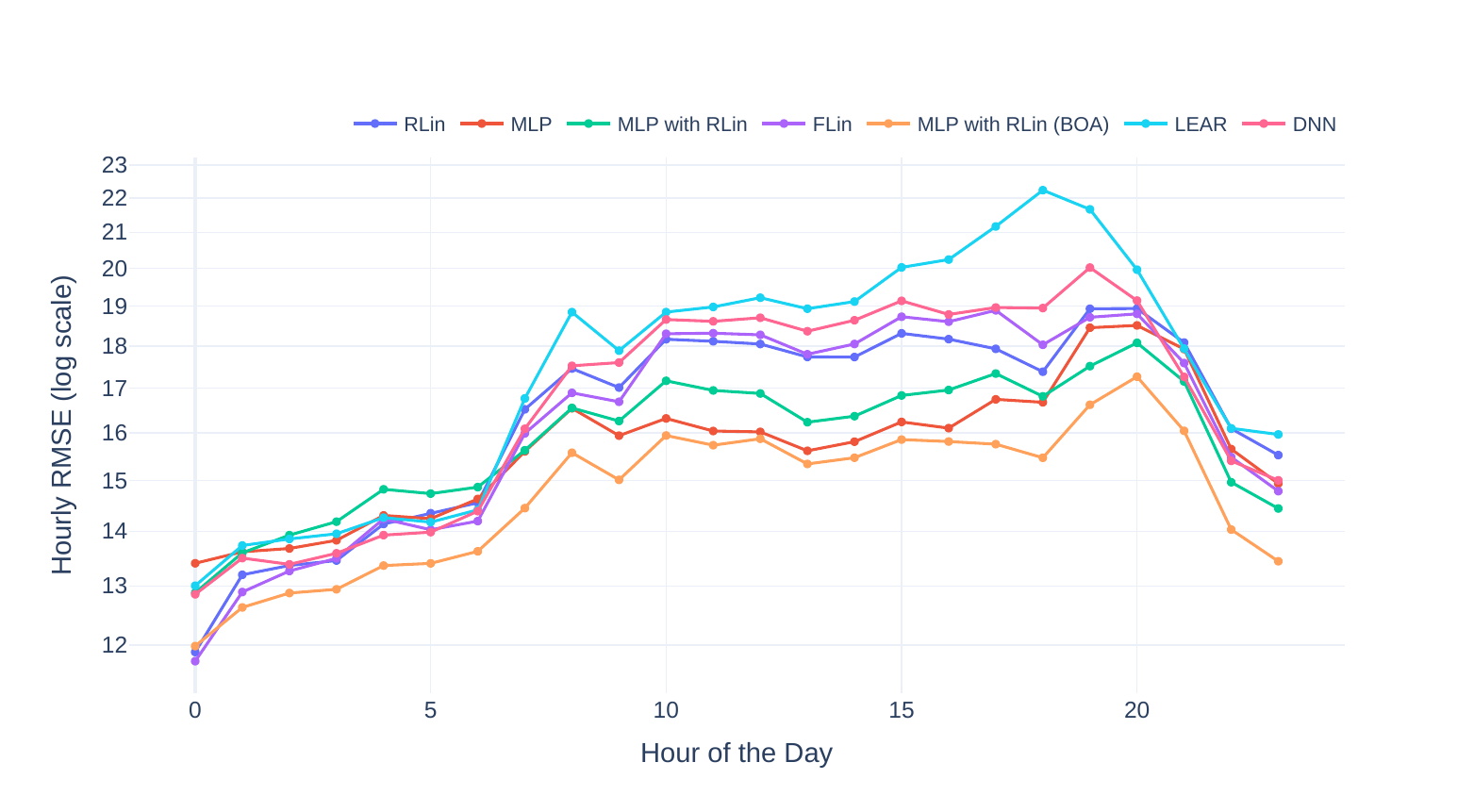}
\caption{Hourly RMSE of selected models for Spanish bidding zone}
\end{figure}

\begin{figure}[h!]
\centering
\includegraphics[width=1.0\linewidth]{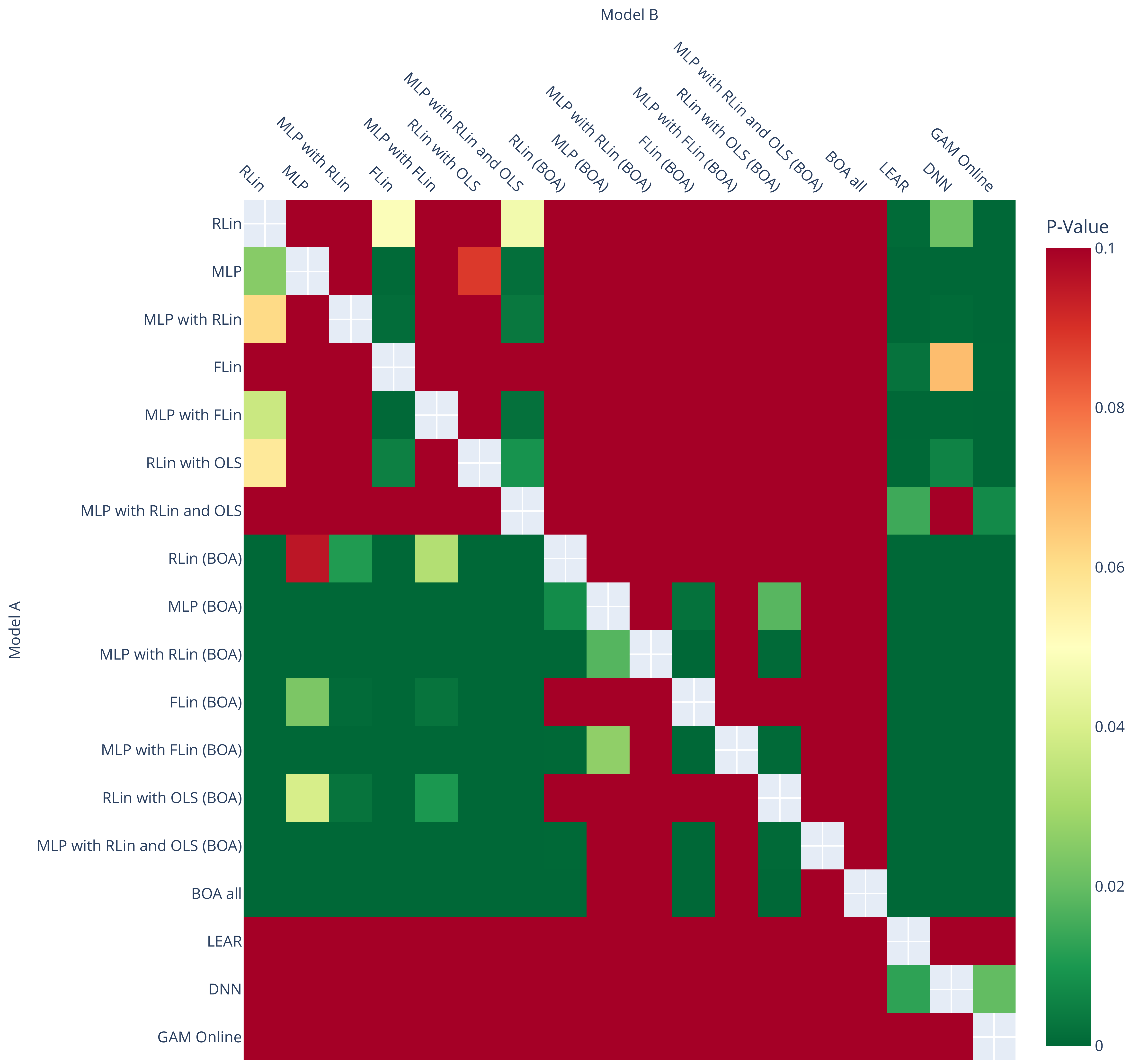}
\caption{Pairwise DM test p-values across forecasting models for Spanish market {using L1 loss function}}
\label{fig16}
\end{figure}

\begin{figure}[h!]
\centering
\includegraphics[width=1.0\linewidth]{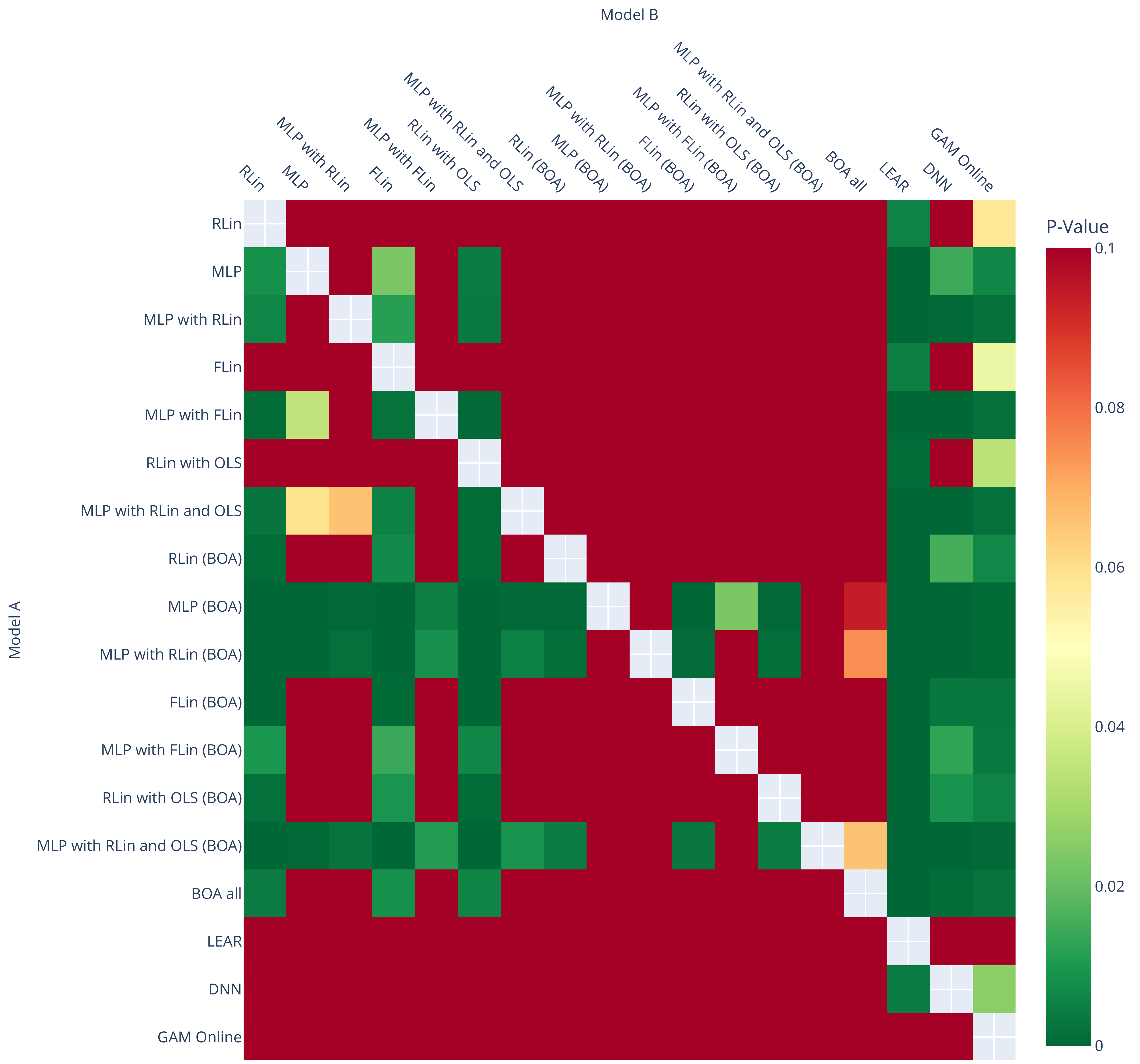}
\caption{{Pairwise DM test p-values across forecasting models for the German-Luxembourg market using L2 loss function}}
\label{fig15}
\end{figure}

\begin{figure}[h!]
\centering
\includegraphics[width=1.0\linewidth]{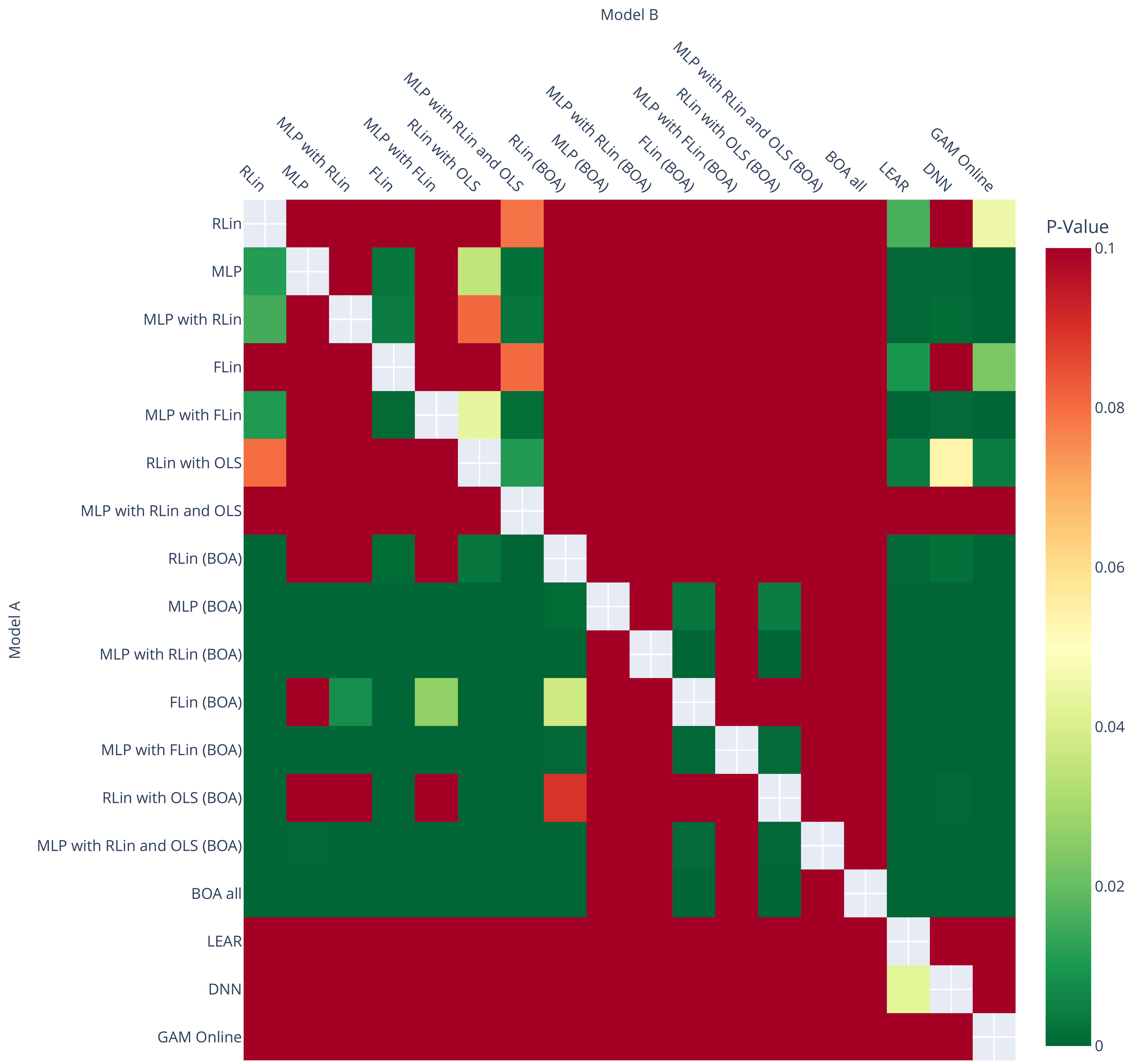}
\caption{{Pairwise DM test p-values across forecasting models for Spanish market using L2 loss function}}
\label{fig17}
\end{figure}

\begin{figure}[h!]
\centering
\includegraphics[width=1.0\linewidth]{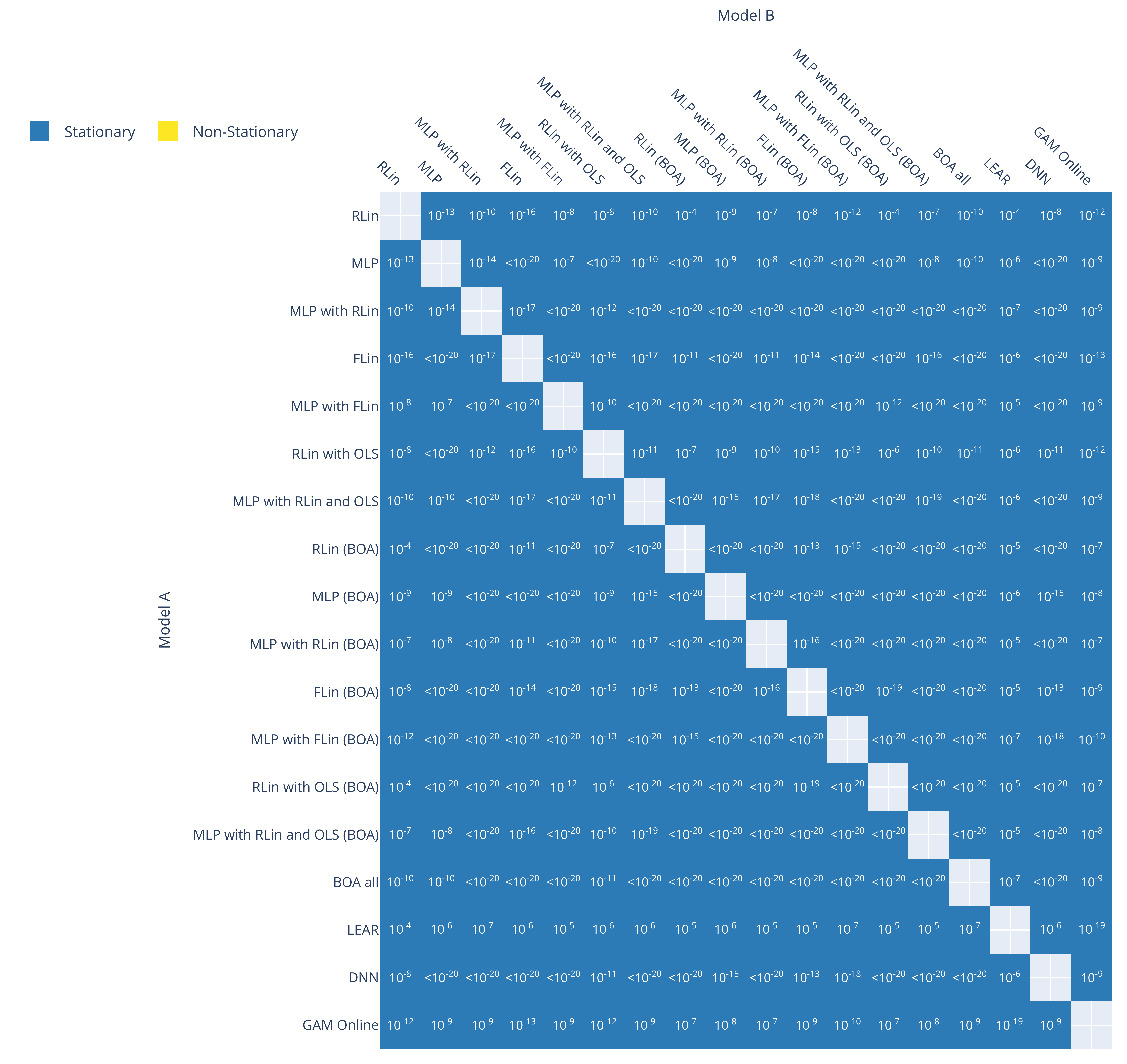}
\caption{{Pairwise stationarity assessment of the L1 loss differential series used in the DM test for German-Luxembourg market, based on the ADF test. Cell entries report the order of magnitude of the ADF p-values. Blue shades indicate rejection of the unit root hypothesis, suggesting stationary loss differentials between the corresponding model pairs, while yellow shades indicate failure to reject the unit root hypothesis. The predominance of blue cells and absence of yellow cells indicate that all pairwise loss-differential series are stationary, supporting the validity of the DM testing framework.}}
\label{sta1}
\end{figure}

\begin{figure}[h!]
\centering
\includegraphics[width=1.0\linewidth]{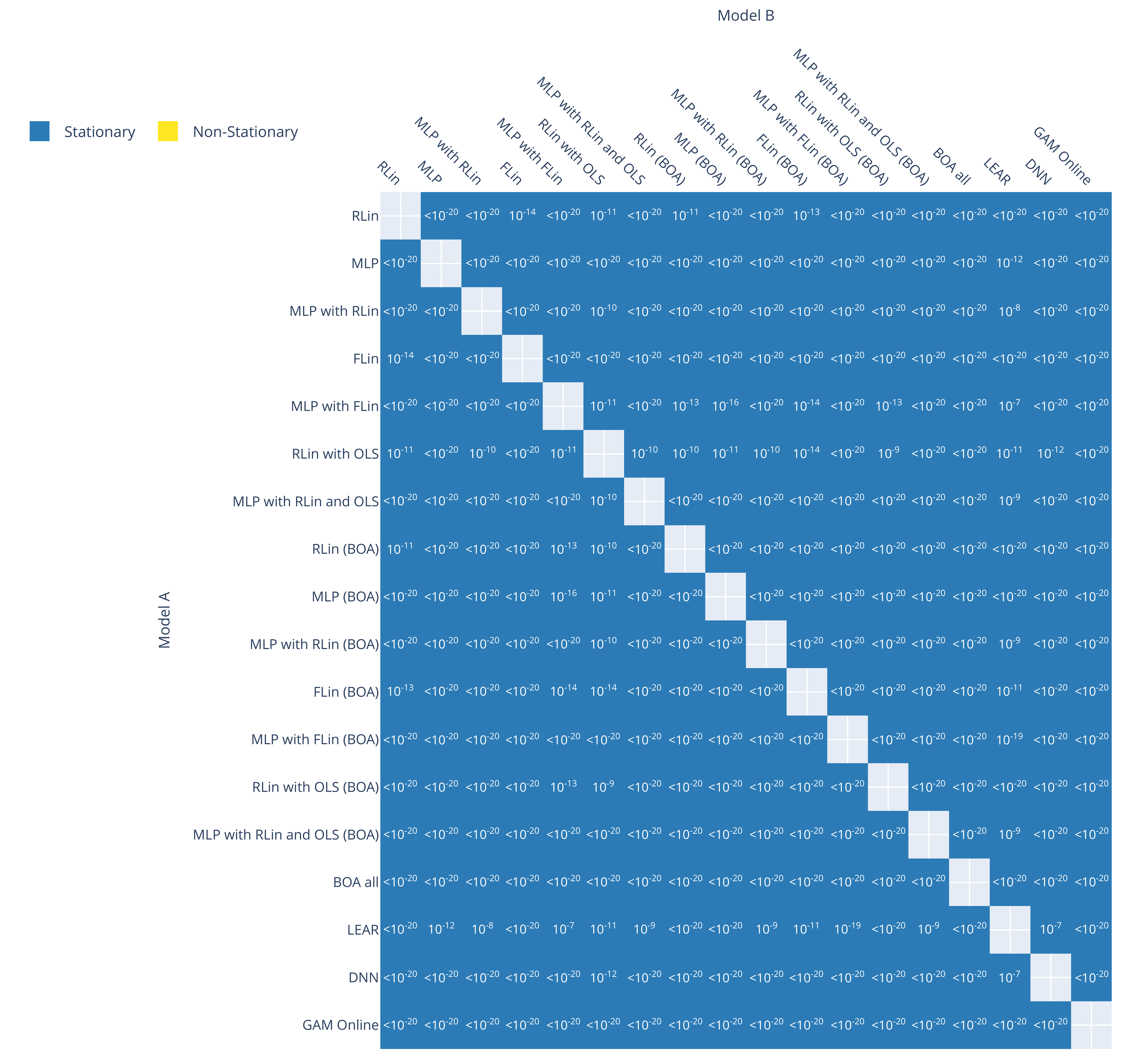}
\caption{{Pairwise stationarity assessment of the L2 loss differential series used in the DM test for German-Luxembourg market, based on the ADF test. Cell entries report the order of magnitude of the ADF p-values. Blue shades indicate rejection of the unit root hypothesis, suggesting stationary loss differentials between the corresponding model pairs, while yellow shades indicate failure to reject the unit root hypothesis. The predominance of blue cells and absence of yellow cells indicate that all pairwise loss-differential series are stationary, supporting the validity of the DM testing framework.}}
\label{sta2}
\end{figure}

\begin{figure}[h!]
\centering
\includegraphics[width=1.0\linewidth]{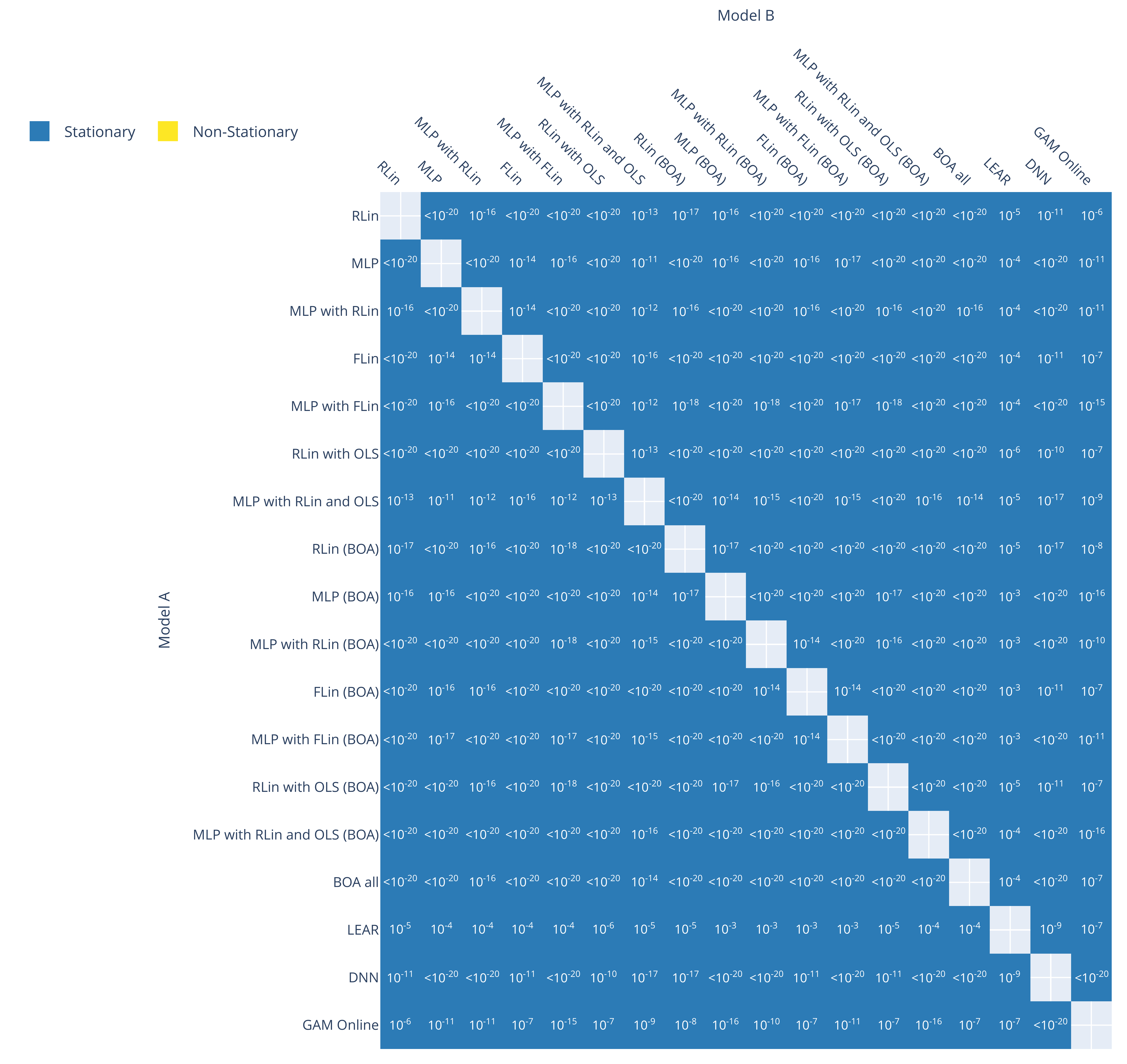}
\caption{{Pairwise stationarity assessment of the L1 loss differential series used in the DM test for Spanish market, based on the ADF test. Cell entries report the order of magnitude of the ADF p-values. Blue shades indicate rejection of the unit root hypothesis, suggesting stationary loss differentials between the corresponding model pairs, while yellow shades indicate failure to reject the unit root hypothesis. The predominance of blue cells and absence of yellow cells indicate that all pairwise loss-differential series are stationary, supporting the validity of the DM testing framework.}}
\label{sta3}
\end{figure}

\begin{figure}[h!]
\centering
\includegraphics[width=1.0\linewidth]{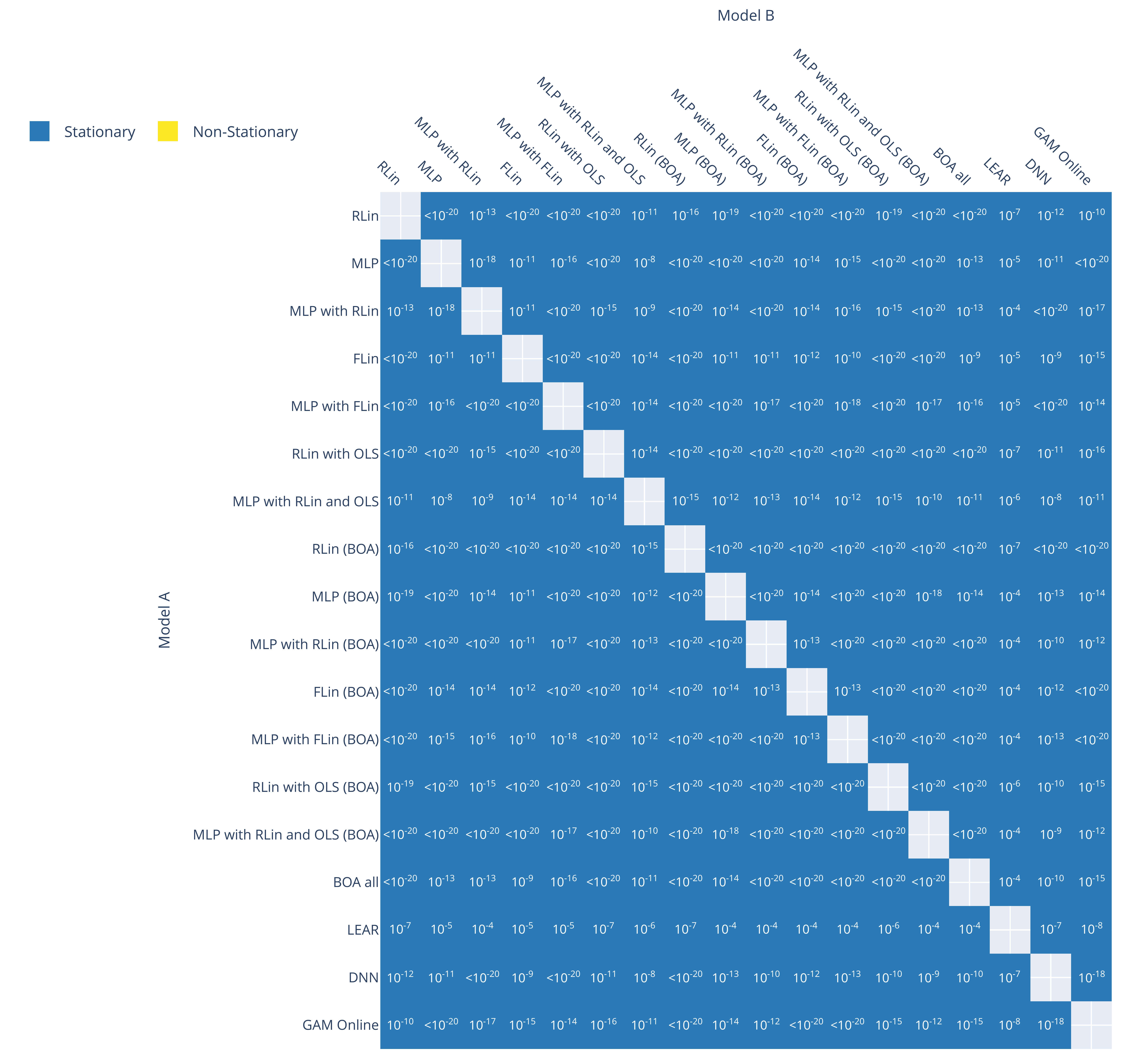}
\caption{{Pairwise stationarity assessment of the L2 loss differential series used in the DM test for Spanish market, based on the ADF test. Cell entries report the order of magnitude of the ADF p-values. Blue shades indicate rejection of the unit root hypothesis, suggesting stationary loss differentials between the corresponding model pairs, while yellow shades indicate failure to reject the unit root hypothesis. The predominance of blue cells and absence of yellow cells indicate that all pairwise loss-differential series are stationary, supporting the validity of the DM testing framework.}}
\label{sta4}
\end{figure}

\begin{table}[h!]
\centering
\begin{tabular}{lccc}
\toprule
Model & MAE & RMSE & Runtime (sec) \\
\midrule
MLP with RLin & \cellcolor{rgb,255:red,212;green,212;blue,255}2.94 & \cellcolor{rgb,255:red,246;green,246;blue,255}5.20 & \cellcolor{rgb,255:red,178;green,178;blue,255}\textbf{63.80} \\
MLP with RLin (BOA) & \cellcolor{rgb,255:red,178;green,178;blue,255}\textbf{2.90} & \cellcolor{rgb,255:red,178;green,178;blue,255}\textbf{5.08} & \cellcolor{rgb,255:red,187;green,187;blue,255}548.40 \\
DNN & \cellcolor{rgb,255:red,255;green,178;blue,178}3.08 & \cellcolor{rgb,255:red,255;green,178;blue,178}5.35 & \cellcolor{rgb,255:red,255;green,178;blue,178}8507.20 \\
\bottomrule
\end{tabular}
\caption{{Comparison of forecasting accuracy and runtime between the proposed model and the benchmark model.  The difference between the DNN results reported in this table and those presented in \textcite{a} arises for two reasons. First, we perform 500 trials for hyperparameter tuning, whereas they use 1500 trials. Second, we report the computational time for the entire two-year forecasting study, while they report the time required for a single training window only.}}
\label{tab_PJM}
\end{table}

\makeatother

\end{document}